\documentclass{article}
\pdfoutput=1

    \PassOptionsToPackage{numbers, sort}{natbib}

\usepackage[final]{neurips_2024}

\usepackage[utf8]{inputenc} 
\usepackage[T1]{fontenc}    
\usepackage{hyperref}       
\usepackage{url}            
\usepackage{booktabs}       
\usepackage{amsfonts}       
\usepackage{nicefrac}       
\usepackage{microtype}      
\usepackage{xcolor}         

\usepackage{amsmath}
\usepackage{amssymb}
\usepackage{graphicx}
\usepackage{tcolorbox}
\usepackage{multirow}
\usepackage{subcaption}
\usepackage{adjustbox}
\usepackage{colortbl}
\usepackage{outlines}
\newcommand{\hjk}[1]{{\color[rgb]{0.0 0.0 0.0}{#1}}}

\newcommand{\Sb}{\mathbf{S}}
\newcommand{\Gb}{\mathbf{G}}
\newcommand{\Tb}{\mathbf{T}}

\newcommand{\yb}{\mathbf{y}}

\newcommand{\mycomment}[1]{}

\title{LLaMo: Large Language Model-based\\Molecular Graph Assistant}

\author{Jinyoung Park \quad Minseong Bae \quad Dohwan Ko \quad Hyunwoo J. Kim\thanks{corresponding author.}\\
        Department of Computer Science and Engineering, Korea University\\
        {\tt \{lpmn678, bms2002, ikodoh, hyunwoojkim\}@korea.ac.kr}}

\begin{document}

\maketitle

\begin{abstract}
    
Large Language Models~(LLMs) have demonstrated remarkable generalization and instruction-following capabilities with instruction tuning.
The advancements in LLMs and instruction tuning have led to the development of Large Vision-Language Models~(LVLMs).
However, the competency of the LLMs and instruction tuning have been less explored in the molecular domain.
Thus, we propose \textcolor{black}{LLaMo: \textbf{L}arge \textbf{La}nguage Model-based \textbf{Mo}lecular graph assistant}, which is an end-to-end trained large molecular graph-language model.
To bridge the discrepancy between the language and graph modalities, we present the multi-level graph projector that transforms graph representations into graph tokens by abstracting the output representations of each GNN layer and motif representations with the cross-attention mechanism.
We also introduce machine-generated molecular graph instruction data to instruction-tune the large molecular graph-language model for general-purpose molecule and language understanding.
Our extensive experiments demonstrate that LLaMo shows the best performance on diverse tasks, such as molecular description generation, property prediction, and IUPAC name prediction.
The code of LLaMo is available at \url{https://github.com/mlvlab/LLaMo}.
\end{abstract}

\section{Introduction}
In recent years, molecular machine learning~\cite{zeng2022deep,taylor2022galactica,liu2023molca,su2022molecular} has received significant attention, addressing diverse tasks in the chemical domain. 
The predominant approach for molecular tasks is graph machine learning~\cite{duvenaud2015convolutional,de2018molgan,atz2021geometric} that leverages the molecular graph structure, which is a natural and expressive representation of molecules.
Although graph-based methods have successfully represented molecules, they have limited interpretability and incompatibility to solve multi-modal molecular tasks dealing with pairs of texts and molecules.
To address these issues, recent works~\cite{edwards2021text2mol,su2022molecular} train both a language model and a graph encoder with cross-modal contrastive learning.
However, the models trained with cross-modal contrastive learning are insufficient to perform open-ended molecule-to-text generation tasks~\cite{liu2023molca}, which are more applicable to practical use. 

Large Language Models~(LLMs)~\cite{taylor2022galactica,touvron2023llama,achiam2023gpt,team2023gemini} have shown impressive progress and accomplished human-like open-ended text generation with the power of billions of parameters.
To leverage the instruction-following capability of LLMs, many works employ instruction-tuning approaches~\cite{wang2023self,chung2024scaling,alpaca} for general-purpose language models.
Motivated by the development of LLMs and instruction tuning, Large Vision-Language Models~(LVLMs) have recently been explored and achieved success on image comprehension and image-to-text generation tasks~\cite{gao2023llama,zhang2023llama,dai2023instructblip,team2023gemini,achiam2023gpt,zhu2024minigpt}.
Despite the success of LLM-based approaches on natural language processing and machine vision domains, the research on the integration of language models and molecular graphs has been less studied due to the lack of consideration of the architecture design of Large Molecular Graph-Language Models~(LMGLMs) and the molecular graph instruction data.

In this paper, we propose LLaMo: \textbf{L}arge \textbf{La}nguage Model-based \textbf{Mo}lecular graph assistant, which seamlessly integrates a molecular graph encoder and a large language model to enable the instruction-following response generation in molecular domain.
Specifically, LLaMo consists of the molecular graph encoder, large language model, and multi-level graph projector that bridges the graph encoder and large language model.
The multi-level graph projector abstracts the representation of each GNN layer and motif representation using a cross-attention mechanism, ensuring a thorough understanding of molecular structures.
Furthermore, we introduce machine-generated molecular graph instruction data through the pipeline to convert molecular descriptions and IUPAC names into a multi-turn conversation format.
The generated instruction-following data enhances the model's ability to perform general-purpose molecule and language understanding, bridging the gap between molecular graph analysis and language-based tasks.
Our proposed LLaMo outperforms the LLM-based works such as GPT-4 across {diverse tasks, including molecular description generation, property prediction, and IUPAC name prediction.

Our contributions are summarized as follows:
\begin{itemize}
    \item We propose LLaMo: \textbf{L}arge \textbf{La}nguage Model-based \textbf{Mo}lecular graph assistant consisting of graph encoder, language model, and multi-level graph projector
    equipped with a multi-level graph projector that captures rich information of the graph structure at multiple levels.
    \item We introduce GPT-4 generated molecular graph-text multi-turn conversation data to address the data scarcity problem of molecule-text datasets and improve the instruction-following capabilities of a large molecular graph-language model.
    \item Our experiments demonstrate that LLaMo achieves the best performance on various tasks such as molecular description generation, property prediction, and IUPAC name prediction.
\end{itemize}
}

\section{Related works}
\noindent \textbf{Molecular graph modeling.}
Molecular graphs serve as a natural and expressive representation of molecules, effectively capturing the structural information. 
Graph neural networks~\cite{kipf2016semi, velickovic2017graph, xu2019powerful,park2022deformable} are commonly utilized architectures for molecular graph representations.
To learn graph neural networks with the limited molecular graph data~\cite{ji2022relmole}, self-supervised learning has been explored.
For example, various approaches~\cite{xia2022mole,zhang2021motif} have been developed to capture multi-level features of molecular graphs, such as node-level masked atom modeling~\cite{xia2022mole}, motif-based self-supervised learning~\cite{zhang2021motif, zang2023hierarchical}, and graph-level contrastive learning~\cite{you2020graph, wang2022molecular}. 
With the advance of multi-modal large language models, molecule-language tasks such as molecule-text retrieval~\cite{edwards2021text2mol} or molecule captioning~\cite{edwards2022translation} have recently drawn significant attention.
Recent works~\cite{zhao2023gimlet,su2022molecular,liu2023molca} have attempted to enable language models to understand molecular graphs. 
\cite{zhao2023gimlet} treated nodes of molecular graphs as tokens of language models. 
Some works have adopted GNN-based encoders, either by propagating their outputs to language models through MLP~\cite{su2022molecular} or employing cross-modal projectors~\cite{liu2023molca}. 
However, these methods fail to consider molecular graphs at multiple levels and are hindered by inherent limitations of graph encoders, such as the over-smoothing problem~\cite{li2018deeper}. 
To address these challenges, we propose a novel architecture, \textcolor{black}{LLaMo}, which effectively propagates multi-level information of molecular graphs to language models.

\noindent \textbf{Instruction tuning.}
Recent advancements of LLMs lead to extensive research on  \emph{instruction tuning}, aimed at improving the model's capability to follow human instructions
~\cite{wei2021finetuned,ouyang2022training,longpre2023flan,sanh2021multitask,wang2023self}. 
To construct high-quality instruction tuning data, a line of previous approaches~\cite{longpre2023flan,sanh2021multitask} has adopted existing human-annotated datasets and integrated them with a new structure and template.
On the other hand, recent studies~\cite{xue2023instruction,vicuna,wang2023self} on instruction tuning have collected data samples from strong LLMs like GPT-4~\cite{achiam2023gpt}.
These works first manually construct annotated seed instruction samples and expand them by prompting LLMs.
As a result, several instruction-tuned LLMs~\cite{vicuna,alpaca,zhang2023llama} have been proposed from the open-source LLMs, \textit{e.g.}, LLaMA~\cite{touvron2023llama} and shown generalizability across a wide range of instructions.
More recently, those studies on instruction tuning have been expanded to visual instruction tuning in image~\cite{gao2023llama,liu2023visual,dai2023instructblip} and video~\cite{maaz2023video,li2023videochat} domain to enable the model to understand the visual contents.
Inspired by the instruction tuning for multi-modal LLMs in other domains, in this work, we study instruction tuning specifically for molecule graphs, which has been underexplored in the literature.
\section{LLaMo: Large Language Model-based Molecular Graph Assistant}
\begin{figure}[t]
    \centering
    \includegraphics[width=1.0\textwidth]{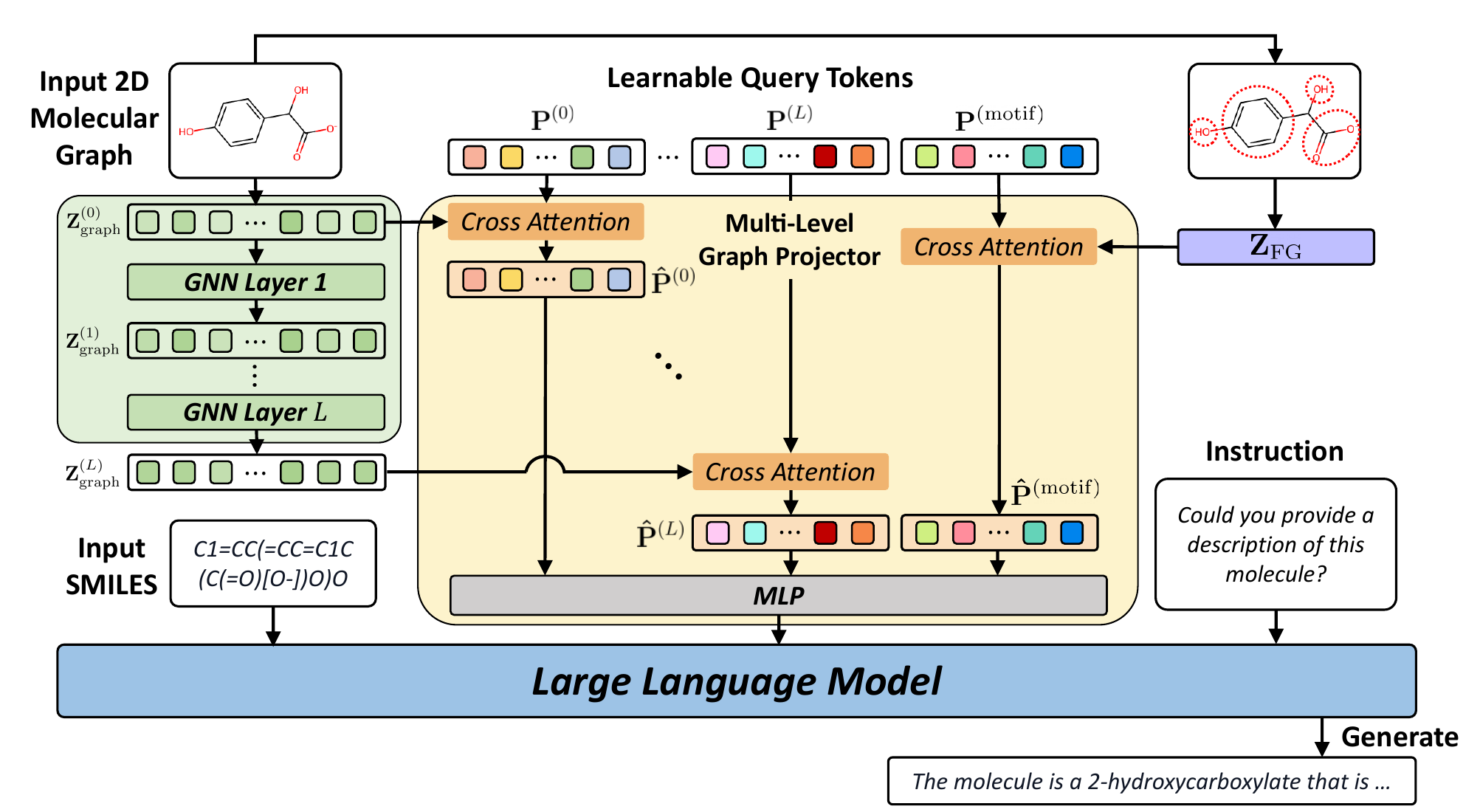}
    \caption{Overall framework of LLaMo. 
    LLaMo consists of a graph neural network, a multi-level graph projector, and a large language model. It first encodes an input 2D molecular graph with the graph neural network and then converts the encoded graph into molecular graph tokens with the multi-level graph projector. Finally, the large language model generates the instruction-following response given the input SMILES, graph tokens, and the instruction.} 
    \label{fig:main_fig}
\end{figure}
The primary goal is to seamlessly integrate a molecular graph encoder and a Large Language Model~(LLM) to generate instruction-following responses to the input texts and molecules.
To achieve it, we propose LLaMo: \textbf{L}arge \textbf{La}nguage Model-based \textbf{Mo}lecular graph assistant, a general-purpose Large Molecular Graph-Language Model~(LMGLM) equipped with a multi-level graph projector.
Specifically, the proposed framework utilizes three input modalities: 1D SMILES~\cite{weininger1988smiles}, 
2D molecular graph, and text~(instruction). 
SMILES~\cite{weininger1988smiles} is a 1D representation of a molecule, and a 2D molecular graph is processed by 
a GNN. 
The three input modalities are fed as a sequence of tokens and our LLaMo autoregressively generates text responses.
Formally, given SMILES $\Sb$, molecular graph tokens $\Gb$, and text (instruction) $\Tb$, 
the proposed method renders the response $\mathbf{Y}=\left\{ \mathbf{y}_i \right \}_{i=1}^K$
as:
\begin{equation}
    p\left(\mathbf{Y} | \mathbf{S}, \mathbf{G}, \mathbf{T} \right) = \prod_{i=1}^K p\left(\mathbf{y}_i | \mathbf{S}, \mathbf{G}, \mathbf{T}, \mathbf{y}_{<i} \right),
\end{equation}
where $\yb_{<i}$ indicates generated token sequences until $i$-th token.

\subsection{Model Architecture}
The overall architecture of \textcolor{black}{LLaMo} is illustrated in Figure~\ref{fig:main_fig}.
\textcolor{black}{LLaMo} consists of a graph encoder, a multi-level graph projector, and a backbone large language model. 
The graph encoder $g(\cdot)$ takes a 2D molecule graph as an input and outputs their node representations as a sequence of tokens.
The multi-level graph projector $\text{Proj}_{\text{MG}}(\cdot)$ transforms the sequence of node representations into molecular tokens to align them with the LLM.
Then, the LLM $f(\cdot)$ processes molecular and text tokens and provides a response in an autoregressive manner.

\noindent\textbf{Graph encoder.} We adopt Graph Neural Networks~(GNNs) as a molecular graph encoder. 
Given the graph $\mathcal{G}$, graph neural networks~$g\left(\cdot\right)$ iteratively update node representation $\mathbf{z}_v^{(l)} \in \mathbb{R}^{d^{(l)}}$ via the message-passing framework.
With the message-passing, $L$-layer GNN provides node representations $\mathbf{z}_v^{(L)}$ that express an $L$-hop ego-graph given the node $v$ as a center node.
\textcolor{black}{More details about graph neural networks are in the Appendix~\ref{app_sec:gnn}.}
\begin{figure}[t]
    \centering
    \includegraphics[width=1.0\textwidth]{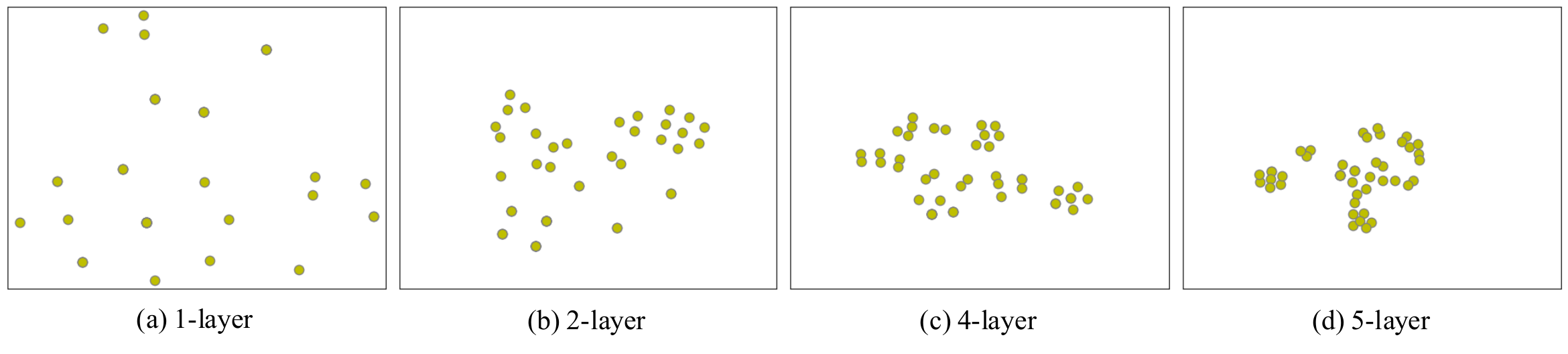}
    \caption{Node representations of graph encoder with 1,2,4,5 layers. As the number of layers increases, node representations collapse.
    } 
    \label{fig:oversmoothing_fig}
\end{figure}

\noindent\textbf{Multi-level graph projector.} The goal of a multi-level graph projector is to align the graph encoder with the LLM by transforming a set of node representations $\mathbf{Z}_{\text{graph}}^{(L)}$ into a sequence of molecular graph tokens $\mathbf{H}_{\text{graph}}$. 
It enables the language model to utilize graph information.
In the literature, projectors have been proposed mainly for Large Vision-Language Models~(LVLMs)~\cite{liu2023visual,ye2023mplug,zhu2024minigpt,dai2023instructblip,cha2024honeybee}.
They are usually implemented using a linear projection~\cite{liu2023visual} or an abstraction of visual features~\cite{ye2023mplug,dai2023instructblip}, which are outputs of the final layer of a visual encoder given input image.
Analogously, we can design the projector for large molecular graph-language models with a linear projection or an abstraction of high-level node representations from the pre-trained graph encoder, which is formulated as:
\begin{equation}
    \mathbf{H}_{\text{graph}}=\text{Proj}\left(\mathbf{Z}_\text{graph}^{(L)} \right), \text{ where }\  \mathbf{Z}_\text{graph}^{(L)} = g\left(\mathcal{G}\right),
\end{equation}
where $\mathbf{Z}_\text{graph}^{(L)}=\left[\mathbf{z}_0^{(L)}, \dots, \mathbf{z}_{\left\vert \mathcal{V}\right\rvert}^{(L)} \right]\in \mathbb{R}^{\left\lvert \mathcal{V}\right\rvert \times {d^{(L)}}}$ is the concatenation of node representation $\mathbf{z}_v^{(L)} \in \mathbb{R}^{d^{(L)}}$ from $L$-th layer GNN and $\text{Proj}\left(\cdot\right)$ is the projector.

However, we observe that the high-level representation is not effective in capturing the local information due to the over-smoothing problem~\cite{li2018deeper}, which means that the node representations become indistinguishable, as the number of layers in the GNN increases.
Figure~\ref{fig:oversmoothing_fig} depicts node representations~(yellow dots) of graph encoder with 1,2,4,5 layers on one molecular graph sample. (\textcolor{black}{More samples are in Appendix~\ref{app_sec:oversmoothing}.})
As mentioned above, node representations become over-smoothed as the number of layers increases, leading to nearly identical node representations in the final layer.
Consequently, conventional projectors relying on high-level node representations have a limited capability to preserve the detailed or local information of molecular graphs. 
Moreover, many tasks require multi-scale information, including atom, atomic group, and molecule levels. 
Hence, the projector that solely utilizes features from the top layer is suboptimal for the tasks.

Motivated by the observations, we propose a novel \textcolor{black}{multi-level graph projector} to generate graph tokens that contain richer information reflecting the graph structure at multiple levels. 
The \textcolor{black}{multi-level graph projector}~ $\text{Proj}_{\text{MG}}\left(\cdot\right)$ is formulated as
\begin{equation}
\mathbf{H}_{\text{graph}}=\text{Proj}_{\text{MG}}\left(\left\{\mathbf{Z}_\text{graph}^{(l)}\right\}_{l=0}^L \right), \text{ where }\  \left\{\mathbf{Z}_\text{graph}^{(l)}\right\}_{l=0}^L = g\left(\mathcal{G}\right).
\end{equation}
The method captures multi-hop graph information by leveraging node representations from all layers of a GNN.  
To handle an arbitrary number of nodes, yielding a variable length $\left\vert \mathcal{V}\right\rvert \times L$ features, 
we adopt the cross-attention with learnable tokens $\mathbf{P}^{(l)}=\left[\mathbf{p}^{(l)}_1, \dots ,\mathbf{p}^{(l)}_b \right] \in \mathbb{R}^{b \times d}$ for $l=0, \dots, L$, where $b$ is the number of learnable prompts.
Here, $[\cdot, \cdot]$ indicates the concatenation operation.
The learnable tokens aggregate $l$-th layer GNN representations into a fixed number of tokens as: 
\begin{equation}
      \hat{\mathbf{P}}^{(l)}= \text{Attn}^{(l)}\left(\mathbf{P}^{(l)}, \mathbf{Z}_\text{graph}^{(l)}, \mathbf{Z}_\text{graph}^{(l)} \right) \in \mathbb{R}^{b \times d},
\end{equation}
where $\text{Attn}\left(Q,K,V\right)$ is the attention operation with query $Q$, key $K$, and value $V$. 

For more detailed representations of the input molecule, LLaMo also has learnable tokens $\mathbf{P}^{(\text{motif})}$ for motif-level representations.
We use the functional groups as motifs, which are the statistically important subgraphs in the molecular graphs.
To construct functional group representations $\mathbf{Z}_{\text{FG}}$, we initially identify functional groups, following \cite{ji2022relmole}.
Then, we vectorize the main characteristics of each functional group, which is represented as $\mathbf{z}_{\text{FG},i}$. 
Finally, the functional group representations~$\mathbf{Z}_{\text{FG}}$ are constructed by concatenating all individual functional group representations~$\mathbf{z}_{\text{FG},i}$, which is formulated as $\mathbf{Z}_{\text{FG}}=\left[\mathbf{z}_{\text{FG},0},\dots,\mathbf{z}_{\text{FG},M}\right]$, where $M$ indicates the number of the functional groups in the given molecular graph.
Given the functional group representations $\mathbf{Z}_{\text{FG}}$ of the input molecule, we obtain $\hat{\mathbf{P}}^{(\text{motif})}= \text{Attn}^{(\text{motif})}\left(\mathbf{P}^{(\text{motif})}, \mathbf{Z}_\text{FG}, \mathbf{Z}_\text{FG} \right)$ with the cross-attention.

Then, we obtain the graph-level representations by applying MLP to the multi-hop and motif-level representations, \textit{i.e.}, $[ \hat{\mathbf{P}}^{0}, \dots, \hat{\mathbf{P}}^{(L)}, \hat{\mathbf{P}}^{(\text{motif})}]$. It is formulated as:
\begin{equation}
      \mathbf{H}_{\text{graph}} = \text{MLP}\left(\left[\hat{\mathbf{P}}^{(0)}, \dots, \hat{\mathbf{P}}^{(L)}, \hat{\mathbf{P}}^{(\text{motif})} \right]\right) \in \mathbb{R}^{b\cdot \left(L+2 \right)\times d},
\end{equation}
where $\mathbf{H}_{\text{graph}}$ is a sequence of graph tokens to be fed into the LLM.
\begin{figure}[t]
    \centering
    \begin{subfigure}[t]{0.47\textwidth}
        \includegraphics[width=\textwidth]{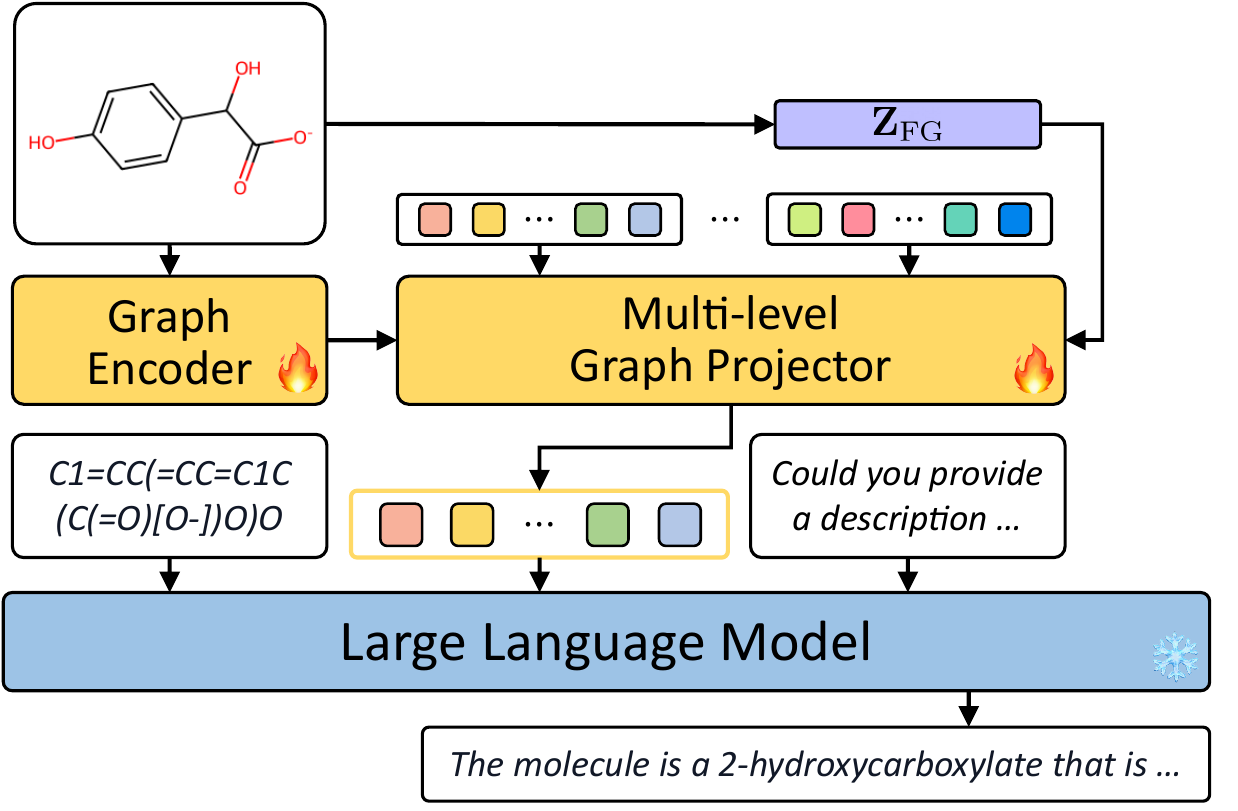}
        \caption{Stage 1: graph-language alignment}
        \label{subfig:stage1}
    \end{subfigure}
    \hfill
    \begin{subfigure}[t]{0.47\textwidth}
        \includegraphics[width=\textwidth]{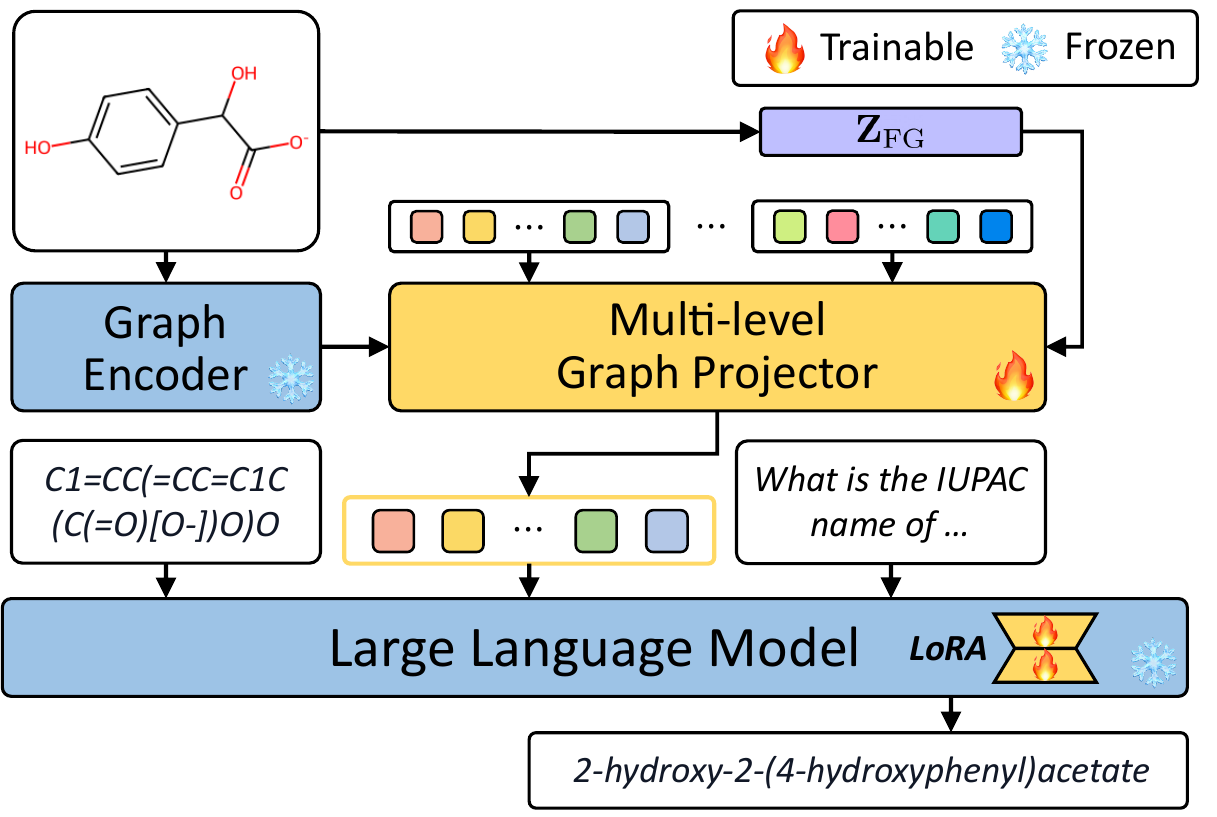}
        \caption{Stage 2: instruction-tuning}
        \label{subfig:stage2}
    \end{subfigure}
    \caption{
    \hjk{
    {Two-stage training pipeline.} 
    Stage 1 involves training the graph encoder, and stage 2 entails fine-tuning the LLM using LoRA. In both stages, the multi-level graph projector is continuously trained. All training processes are performed by generating the instruction-following response.}
    }
    \label{fig:training}
\end{figure}

\noindent\textbf{Large language models.} After constructing tokens for encoding molecular graphs, LLaMo fuses SMILES representation, graph, and text tokens and puts them into the language model to generate an instruction-following response.
\subsection{Training LLaMo}
Similar to most LVLMs~\cite{liu2023visual,ye2023mplug,zhu2024minigpt}, we train LLaMo in the two-stage pipeline: (1) pre-training for molecular graph-language alignment and (2) instruction-tuning end-to-end as in Figure~\ref{fig:training}.

\noindent\textbf{Stage 1. Pre-training for molecular graph-language alignment.}
The first stage focuses on the alignment between the graph encoder and a large language model by learning our multi-level graph projector.
In this stage, with the LLM frozen, we train the multi-level graph projector and the graph encoder by generating molecule descriptions.
For training, we use a molecule-description pair dataset~(\textit{e.g.,} PubChem~\cite{kim2021pubchem}) consisting of a 1D SMILES representation of molecule and molecular graph and its corresponding description.

\noindent\textbf{Stage 2. Instruction-tuning end-to-end.}
In the second stage, we train the LLM to enhance the instruction-following capabilities and enable a deeper understanding of molecular graphs.
In this stage, we freeze the graph encoder and train both the multi-level graph projector and the LLM.
Since it is too expensive to train the full LLM, we employ LoRA~\cite{hu2022lora} to adapt LLM to the data.
For instruction-following, we use the GPT-generated instruction-following multi-turn conversation dataset, which will be introduced in Section~\ref{sec:gpt_instruction}.
In addition to our generated instruction-following dataset, we use a diverse set of datasets with various instructions: molecule description generation, molecular property prediction, IUPAC name generation, forward reaction prediction, and retrosynthesis datasets.
\section{GPT-assisted Molecular Graph Instruction Data Generation}
\label{sec:gpt_instruction}
\begin{table*}[t]
\caption{\textbf{One example to illustrate the instruction-following data.} The top block shows the contexts such as SMILES, description, and IUPAC name used to prompt GPT, and the bottom block shows the response of GPT-4.}
\label{tab:example_conv}
\centering
\begin{minipage}{1.0\columnwidth}\vspace{0mm}    \centering
\begin{tcolorbox} 
    \centering
      \small
    \begin{tabular}{p{0.97\columnwidth} c}
    {\bf Context type 1: SMILES representation}  & \\
CCCCC(C)/C=C(\textbackslash C)/C=C/C(=O)NC1=C[C@]([C@@H](CC1=O)O)(/C=C/C=C/C=C/C(=O)NC2 & \\ 
=C(CCC2=O)O)O & \\

    {\bf Context type 2: Description}  & \\
The molecule is a polyene antibiotic that is TMC-1A in which the 2,4-dimethyloct-2-enoyl group has been replaced by an (E,E)-4,6-dimethyldeca-2,4-dienoyl group. TMC-1C is an antitumour antibiotic isolated from Streptomyces sp. A-230...<omitted>

    {\bf Context type 3: IUPAC name}  & \\
(2E,4E)-N-[(3S,4R)-3,4-dihydroxy-3-[(1E,3E,5E)-7-[(2-hydroxy-5-oxocyclopenten-1-yl)amino]-7-oxohepta-1,3,5-trienyl]-6-oxocyclohexen-1-yl]-4,6-dimethyldeca-2,4-dienamide & \\

    \hrulefill & \\
    {\bf Response: Multi-turn conversation}  & \\
Question: &\\
What is the IUPAC name of the molecule you are analyzing? &\\
=== &\\
Answer: &\\
The IUPAC name of the molecule is (2E,4E)-N-[(3S,4R)-3,4-dihydroxy-3-[(1E,3E,5E)-7-[(2-hydroxy-5-oxocyclopenten-1-yl)amino]-7-oxohepta-1,3,5-trienyl]-6-oxocyclohexen-1-yl]-4,6-dimethyldeca-2,4-dienamide. &\\
=== &\\
Question: &\\
Can you identify the type of acid or base this molecule can act as in a reaction? &\\
=== &\\
Answer: &\\
This molecule can act as a weak acid due to the presence of the carboxylic acid group. &
    \end{tabular}
\end{tcolorbox}

\end{minipage}
\end{table*}

Instruction data are essential for improving the instruction-following capabilities of LLM-based models. 
Despite active research on instruction-tuning, the instruction-following data for molecular graphs have been less explored in the literature since annotations require expertise. 
To alleviate the need for expertise and minimize the manual efforts,
we utilize GPT-4~\cite{achiam2023gpt} to generate molecular graph-text instruction-following data using graph-text pair datasets.

Inspired by previous results~\cite{xu2023lvlm,li2023textbind}, we construct multi-turn conversation datasets, which are more diverse and effective for instruction tuning compared to simple pairs of questions and answers.
We leverage GPT-4 to generate multi-turn conversations with tailored contexts/prompts that consist of two representations for molecular graphs and description: (i) SMILES representation that describes the chemical structures with special strings, (ii) captions that explain the molecule, and (iii) IUPAC name that describes the molecule based on its chemical composition and structure.
These representations enable the GPT-4, which inherently lacks in-depth molecular knowledge, to understand and generate a diverse and high-quality set of examples.
One example of the input representations is shown in the top block of Table~\ref{tab:example_conv}.

Specifically, we generate the multi-turn conversation data in three steps: 
\textbf{1)} select exemplar conversations among machine-generated instruction-tuning data, \textbf{2)} generate multi-turn conversations via in-context learning with the exemplar conversations as prompts, and \textbf{3)} filter out incomplete conversations and those with many turns.
In the first step, we generate exemplars with a brief human-written instruction as shown in Appendix~\ref{app_sec:prompts}. 
However, we found that GPT-4 frequently fails to generate complete multi-turn conversations without the exemplars.
To address this issue, we generate the instruction data with in-context learning.
We sample exemplars from a small set of complete conversations generated by GPT-4 in the first step.
Then, GPT-4 generates the complete multi-turn conversation data for the instruction tuning guided by the prompts wrapped with the generated exemplars.
To validate the quality of the generated conversation, we sample 500 subsets generated via in-context learning.
We find that some conversations consisting of a large number of turns are prone to generating incomplete and inaccurate outputs.
So, we filter out incomplete conversations and those with many turns.
The example of the generated multi-turn conversation is in the bottom block of Table~\ref{tab:example_conv}.
In total, we generate 12K unique molecular graph-language instruction-following samples using PubChem324k dataset~\cite{liu2023molca,kim2021pubchem}.

\section{Experiments}
\subsection{Experimental Settings}

\begin{table*}[!t]
    \caption{\hjk{Performance (\%) of generalist models on three tasks: molecule description generation, IUPAC prediction, and property prediction.
    \textbf{Mol. Inst. tuned} denotes the molecular instruction-tuned model.
    $*$ The result is not available since LLaMA2 fails generating numerical outputs.
    $\dagger$ denotes the experimental results drawn from Mol-Instruction~\cite{fang2024mol}. 
    }}
    \label{tab:generalist}
    \centering
    \begin{adjustbox}{width=\textwidth}
    \begin{tabular}{c c c| c c|c c|c}
        \toprule
        \multirow{2}{*}{\textbf{Model}} & \multirow{2}{*}{\textbf{LLM}} &  \textbf{Mol. Inst.}&\multicolumn{2}{c|}{\textbf{Molecule Description}} & \multicolumn{2}{c|}{\textbf{IUPAC Prediction}} & \textbf{Property pred.} \\
         & &  \textbf{tuned}&BLEU~($\uparrow$) & METEOR~($\uparrow$) & BLEU~($\uparrow$) & METEOR~($\uparrow$) & MAE~($\downarrow$) \\
        \midrule
        \midrule
         \multicolumn{1}{l}{GPT-3.5} & GPT-3.5  && 2.2&19.7 &33.4 &52.6 &0.075\\
         \multicolumn{1}{l}{GPT-3.5 (ICL)} & GPT-3.5 & & 28.4&56.1& 50.3& 62.0&0.028\\
         \multicolumn{1}{l}{GPT-4} & GPT-4  && 0.8 & 16.7&29.0 &48.1 &0.098\\
         \multicolumn{1}{l}{GPT-4 (ICL)} & GPT-4  && 27.0& 52.2&51.8 &62.4 &0.019\\
         \multicolumn{1}{l}{Galactica$\dagger$} & Galactica  && 0.8& 6.5& --&-- &0.568\\
         \multicolumn{1}{l}{Text+Chem T5$\dagger$} & T5-Base && 3.6  & 13.9 & --& --& --  \\
         \midrule
         \multicolumn{1}{l}{LLaMA2} & LLaMA2-7B  && 0.0 & 14.1& 0.0& 0.4&N/A$^*$\\
         \multicolumn{1}{l}{Mol-Instructions$\dagger$} & LLaMA2-7B  &\checkmark& 14.3 & 25.4 & -- & -- &0.013\\
        \cmidrule{1-8}
         \multicolumn{1}{l}{\textbf{LLaMo (Ours)}} & LLaMA2-7B & \checkmark& \textbf{38.9} & \textbf{67.1} & \textbf{56.0}&\textbf{73.2}&\textbf{0.006}\\
        \bottomrule
    \end{tabular}
\end{adjustbox}
\end{table*}

\noindent\textbf{Benchmarks.}
To evaluate the efficacy of the proposed method, we evaluate the model for three tasks such as \textbf{1)} molecule description generation, \textbf{2)} IUPAC name prediction, \textbf{3)} property prediction~(regression).
We conducted experiments under two major settings: generalist and specialist models.
In the generalist setting, one model handles all three tasks, whereas in the specialist setting, we train a model for each downstream task.
More details about benchmarks are in Appendix~\ref{sup:datasets}.

\noindent\textbf{Implementation details.}
For the generalist models, we train our LLaMo based on \texttt{Llama-2-7b-chat}~\cite{touvron2023llama} for a fair comparison with Mol-Instructions~\cite{fang2024mol}.
For the specialist models, we train our LLaMo with Galactica 1.3B~\cite{taylor2022galactica} for a fair comparison with MolCA~\cite{liu2023molca}.
To train the generalist variant of LLaMo, we use a training split of molecular description generation dataset of Mol-Instruction~\cite{fang2024mol} in stage 1.
In stage 2, the model is instruction-tuned with a training split of description generation, property prediction, forward reaction, and retrosynthesis instruction dataset of Mol-Instruction~\cite{fang2024mol}, IUPAC name prediction from \cite{liu2023molca}, and our {GPT-generated} instruction-following data.
To train the specialist variant of LLaMo, we follow MolCA~\cite{liu2023molca} to train the model with a pretraining split of PubChem324kV2 in the stage 1 phase and fine-tune the model for each specific downstream task in the stage 2.
We adopt a long training schedule~(epoch 1 pre-training, epoch 3 instruction tuning) for the final models.
For analysis, we use a short training schedule~(epoch 1 pre-training, epoch 1 instruction tuning).
For further implementation details, refer to Appendix~\ref{sup:implementation_details}.

\noindent\textbf{Baselines.}
For the generalist models, we compare our LLaMo with (1) LLM-based generalist models including Galactica~\cite{taylor2022galactica}, {LLaMA2-7B}~\cite{touvron2023llama}, GPT-3.5, and GPT-4,
(2) Molecule-specialized LLM such as Text+Chem T5~\cite{christofidellis2023unifying}, and
(3) Molecule instruction-tuned generalist model such as Mol-Instructions~\cite{fang2024mol}.
Since GPT-3.5 and GPT-4 have difficulty in solving the tasks without in-context learning, we additionally measure the performance of GPT-3.5 and GPT-4 with 4-shot in-context learning, which are GPT-3.5 (ICL) and GPT-4 (ICL).
For the specialist models, we use single-task specialist molecule-language models as baselines, including MolT5~\cite{edwards2022translation}, MoMu~\cite{su2022molecular}, and MolCA~\cite{liu2023molca}.

\subsection{Experimental Results}
\begin{table*}[!t]
    \centering
    \caption{Performance (\%) of specialist models on molecule captioning with the PubChem324k and ChEBI-20 datasets and IUPAC name prediction. Full ft denotes full parameter fine-tuning.}
    \label{tab:specialist}
    \begin{adjustbox}{width=1.0\textwidth}
    \begin{tabular}{c c c|c c|c c|c}
        \toprule
         \multirow{2}{*}{\textbf{Model}} & \multirow{2}{*}{\textbf{LLM}} & \multirow{2}{*}{\textbf{Training type}} & \multicolumn{2}{c|}{\textbf{PubChem324kV2}} & \multicolumn{2}{c|}{\textbf{ChEBI-20}} & \textbf{IUPAC} \\
        & & & BLEU & METEOR & BLEU & METEOR & METEOR \\
        \midrule
        \midrule
        \multicolumn{1}{l}{MolT5-Small} & T5-Small & full ft & 8.5 & 18.5 & 43.6 & 55.1  & 42.5\\
        \multicolumn{1}{l}{MolT5-Base} & T5-Base & full ft & 20.9 & 35.6 & 45.7 & 56.9 & 53.2\\
        \multicolumn{1}{l}{MolT5-Large} & T5-Large & full ft & 22.2 & 36.6 & 50.8 & 61.4 & 58.5 \\
        \cmidrule{1-8}
        \multicolumn{1}{l}{MoMu-Small} & T5-Small &  full ft & 12.0 & 21.8 & 44.5 & 57.6 & -- \\
        \multicolumn{1}{l}{MoMu-Base} & T5-Base & full ft & 21.5 & 34.2 & 46.2 & 57.6 & --\\
        \multicolumn{1}{l}{MoMu-Large} & T5-Large & full ft & 22.8 & 36.2 & 51.5 & 59.7 & --\\
        \multicolumn{1}{l}{MolCA, $\text{Galac}_{\text{125M}}$} & Galactica-125M & full ft & 24.3 & 41.6 & 52.6 & 63.6& 71.8 \\
        \multicolumn{1}{l}{MolCA, $\text{Galac}_{\text{1.3B}}$} & Galactica-1.3B & LoRA & 30.3 & 45.6 & 53.1 & 65.1 & 72.1\\
        \cmidrule{1-8}
        \multicolumn{1}{l}{\textbf{LLaMo (Ours)}} & Galactica-1.3B &  LoRA & \textbf{34.4} & \textbf{48.0} & \textbf{54.8}& \textbf{66.6}&\textbf{73.4} \\
        \bottomrule
    \end{tabular}
    
\end{adjustbox}
\end{table*}

\noindent\textbf{Generalist models.}
We provide the experimental results of generalist models in molecular description generation, IUPAC name generation, and property prediction tasks.
Our LLaMo is built on LLaMA-7B and it is fine-tuned by our instruction-tuning method.
Table~\ref{tab:generalist} shows that our LLaMo achieves the best performance in all three tasks.
In comparison to \textbf{GPT-4 (ICL)}, which is GPT-4 with in-context-learning, 
LLaMo shows a performance improvement of 11.9 in BLEU-4 and 14.9 in METEOR for molecular description generation.
Furthermore, LLaMo outperforms Mol-Instructions, an instruction-tuned model with molecular data, by a substantial performance gain of 41.7 in METEOR for molecular description generation and a 0.007 performance gain in MAE on the property prediction task.
More experimental results on forward reaction prediction and retrosynthesis are in Appendix~\ref{app_sec:add}.

\noindent\textbf{Specialist models.}
We also evaluate the performance of specialist models to validate the effectiveness of our LLaMo, which is individually fine-tuned for each dataset.
Table~\ref{tab:specialist} demonstrates that our LLaMo consistently achieves the best performance across all tasks and datasets.
Specifically, LLaMo outperforms the second-best model MolCA with Galactica 1.3B, by 4.1 in BLEU-score and 2.4 in METEOR on the PubChem324kV2 dataset.
For IUPAC name prediction, LLaMo also shows superior performance, achieving a METEOR score of 73.4, which surpasses MolCA with Galactica 1.3B by a margin of 1.3 points.
This experimental result indicates that our LLaMo is consistently effective in comprehending molecular graphs based on diverse large language models.

\subsection{Analysis}

\begin{table*}[!t]
    \centering
    \caption{Performance comparison according to the projector type.}
    \label{tab:abl}
    \begin{tabular}{c |c  c|c c|c}
        \toprule
        \multirow{2}{*}{\textbf{Projector}} & \multicolumn{2}{c|}{\textbf{Molecule description}} & \multicolumn{2}{c|}{\textbf{IUPAC prediction}} & \textbf{Property QA} \\
         & BLEU~($\uparrow$)  & METEOR~($\uparrow$) & BLEU~($\uparrow$) & METEOR~($\uparrow$) & MAE~($\downarrow$) \\
        \midrule
        \midrule
        w/o Graph & 26.1   & 56.6 & 36.3  & 62.2 & 0.013\\
         MLP (w/ low-level)   & 32.4&62.1 & 42.2& 68.4& 0.009\\
         MLP (w/ high-level)  & 33.8&63.4 & 45.5& 67.4& 0.008\\
         MLP (w/ concat)  & 34.8 &64.1 &47.1 & 70.2& \textbf{0.007}\\
         Resampler  & 34.4 &62.8 & 43.4& 65.2 &0.009\\
         MGProj (w/o motif)& 36.1& 65.3& 48.8& 69.8& 0.008\\
        \midrule
         {\textbf{MGProj (Ours)}}  & \textbf{37.8}  & \textbf{66.1} & \textbf{49.6}& \textbf{70.9}& \textbf{0.007}\\
        \bottomrule
    \end{tabular}
\end{table*}

\noindent\textbf{Impact of multi-level graph projector.}
To validate the effectiveness of our multi-level graph projector, 
we compare the performance of the multi-level graph projectors (denoted by \textbf{MGProj}) with other projectors in Table~\ref{tab:abl}, including two widely-used projectors such as MLPs and resamplers.
Additionally, we measure the performance of the base model without a graph (and a projector) denoted as \textbf{w/o Graph} for the ablation study.
\textbf{MLP~(w/ low-level)} and \textbf{MLP~(w/ high-level)} denote the MLP projectors where the input is low-level representation~$\mathbf{Z}^{(1)}_{\text{graph}}$ and high-level representation~$\mathbf{Z}^{(L)}_{\text{graph}}$, respectively.
\textbf{MLP~(w/ concat)} indicates the MLP projector with the concatenated representations of all GNN layers as an input.
\textbf{Resampler} denotes the cross-attention based resampler projector designed in Qwen-VL~\cite{bai2023qwen}.
\textbf{MGProj (w/o motif)} and \textbf{MGProj} are our multi-level graph projector without and with motif tokens~$\mathbf{\hat{P}}^{\text{(motif)}}$.

Table~\ref{tab:abl} shows that our multi-level graph projector~(\textbf{MGProj}) achieves the best performance across all three tasks.
Specifically, the multi-level graph projector achieves 49.6 BLEU and 70.9 METEOR scores with a significant improvement compared to MLP projectors in the IUPAC prediction task.
These experimental results demonstrate that our multi-level graph projector is more effective than conventional projectors by capturing multi-scale information, including atom, atomic group, and molecule-level information.

\begin{table*}[!ht]
    \centering
    \caption{Ablation studies on training stage and GPT-generated instruction tuning data.}
    \label{tab:abl2}
    \begin{adjustbox}{width=1.0\textwidth}
    \begin{tabular}{ccc|cc|cc|c}
    \toprule
         \multirow{2}{*}{\textbf{Stage 1}}& \multirow{2}{*}{\textbf{Stage 2}}& \multirow{2}{*}{\textbf{GPT-generated data}}&\multicolumn{2}{c|}{\textbf{Molecule description}} & \multicolumn{2}{c|}{\textbf{IUPAC prediction}} & \textbf{Property QA}  \\
         &&& BLEU~($\uparrow$)  & METEOR~($\uparrow$) & BLEU~($\uparrow$) & METEOR~($\uparrow$) & MAE~($\downarrow$) \\
         \midrule
         \midrule
         &&& 0.0&14.1 &0.0 &0.4 & N/A \\
         \checkmark &&&35.5 &64.8 & 7.3 & 16.9 & N/A\\
         \checkmark & \checkmark &&37.2 &65.1 & 47.5 & 70.2 & \textbf{0.007}\\
         \checkmark & \checkmark & \checkmark & \textbf{37.8}  & \textbf{66.1} & \textbf{49.6}& \textbf{70.9}& \textbf{0.007}\\
    \bottomrule
    \end{tabular}
    \end{adjustbox}
\end{table*}

\begin{table*}[!ht]
    \centering
    \caption{Performance comparison according to the training type.}
    \label{tab:inst_multitask}
    \begin{adjustbox}{width=1.0\textwidth}
    \begin{tabular}{c|cc|cc|c}
    \toprule
         \multirow{2}{*}{\textbf{Training type}}& \multicolumn{2}{c|}{\textbf{Molecule description}} & \multicolumn{2}{c|}{\textbf{IUPAC prediction}} & \textbf{Property QA}  \\
         & BLEU~($\uparrow$)  & METEOR~($\uparrow$) & BLEU~($\uparrow$) & METEOR~($\uparrow$) & MAE~($\downarrow$) \\
         \midrule
         \midrule
         w/o inst. tuning~(Stage 1)  &35.5 &64.8 & 7.3 & 16.9 & N/A\\
         Multi-task & 36.9 & 64.2 & 49.4 & 70.5 & 0.218  \\
         \textbf{Instruction-tuning~(Ours)}  & \textbf{37.8}  & \textbf{66.1} & \textbf{49.6}& \textbf{70.9}& \textbf{0.007}\\
    \bottomrule
    \end{tabular}
    \end{adjustbox}
\end{table*}
\begin{figure}[!ht]
    \centering
    \includegraphics[width=1.0\textwidth]{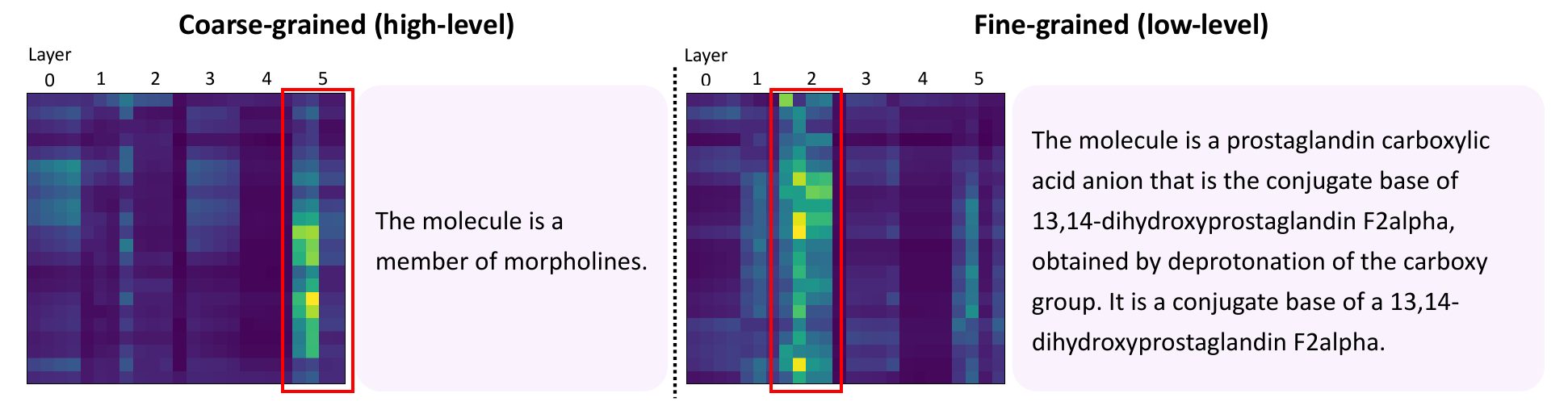}
    \caption{
    \hjk{Visualization of attention maps for samples with coarse-grained caption~(left) and fine-grained caption~(right). 
    The attention scores of high-level features are relatively high when generating coarse-grained captions, whereas those of low-level features are high for fine-grained captions.}
    } 
    \label{fig:attmap}
\end{figure}

\noindent\textbf{Impact of \textcolor{black}{GPT-generated} instruction-tuning data.}
In Table~\ref{tab:abl2}, we provide the ablation studies of each training stage and our GPT-generated instruction dataset.
The experimental results reveal that the instruction tuning with our generated multi-turn conversation data enhances the performance of LLaMo compared to the models trained via one or two-stage training without our GPT-generated instruction data.
This indicates that instruction tuning with our GPT-generated multi-turn conversation data provides the model with more detailed and instruction-following guidance.

\noindent\textbf{Instruction tuning v.s. multi-task learning.}
Table~\ref{tab:inst_multitask} shows the advantages of instruction-tuning based on task instructions compared to multi-task learning using the simple task identifier. 
We use the task name as a simple task identifier for multi-task learning.
From the table, the model without instruction tuning (Stage 1) achieves BLUE score of 35.5 and 7.3 on molecule description and IUPAC prediction tasks, respectively. 
The multi-task learning approach improves the scores to 36.9 for molecule description and 49.4 for IUPAC prediction. 
However, the instruction-tuning method demonstrated the most significant enhancement, achieving the highest scores of 37.8 for molecule description and 49.6 for IUPAC prediction. 
These results indicate that instruction tuning outperforms both the baseline and multi-task learning methods, suggesting its effectiveness in improving model performance on general-purpose training.

\noindent\textbf{Visualization of attention maps.}
We visualize the attention map to explore the effect of the multi-level graph projector in Figure~\ref{fig:attmap}.
The figure illustrates the attention maps of graph tokens for generating coarse-grained~(left) and fine-grained~(right) descriptions.
Interestingly, the attention scores of the low-level are relatively higher than the high-level when generating fine-grained captions, whereas the attention value of the high levels is high when generating coarse-grained captions.
This indicates that both low and high-level graph structural information is crucial in expressing the molecules, and the attention matrix is adaptive to the caption types. 

\begin{figure}[t]
    \centering
    \includegraphics[width=1.0\textwidth]{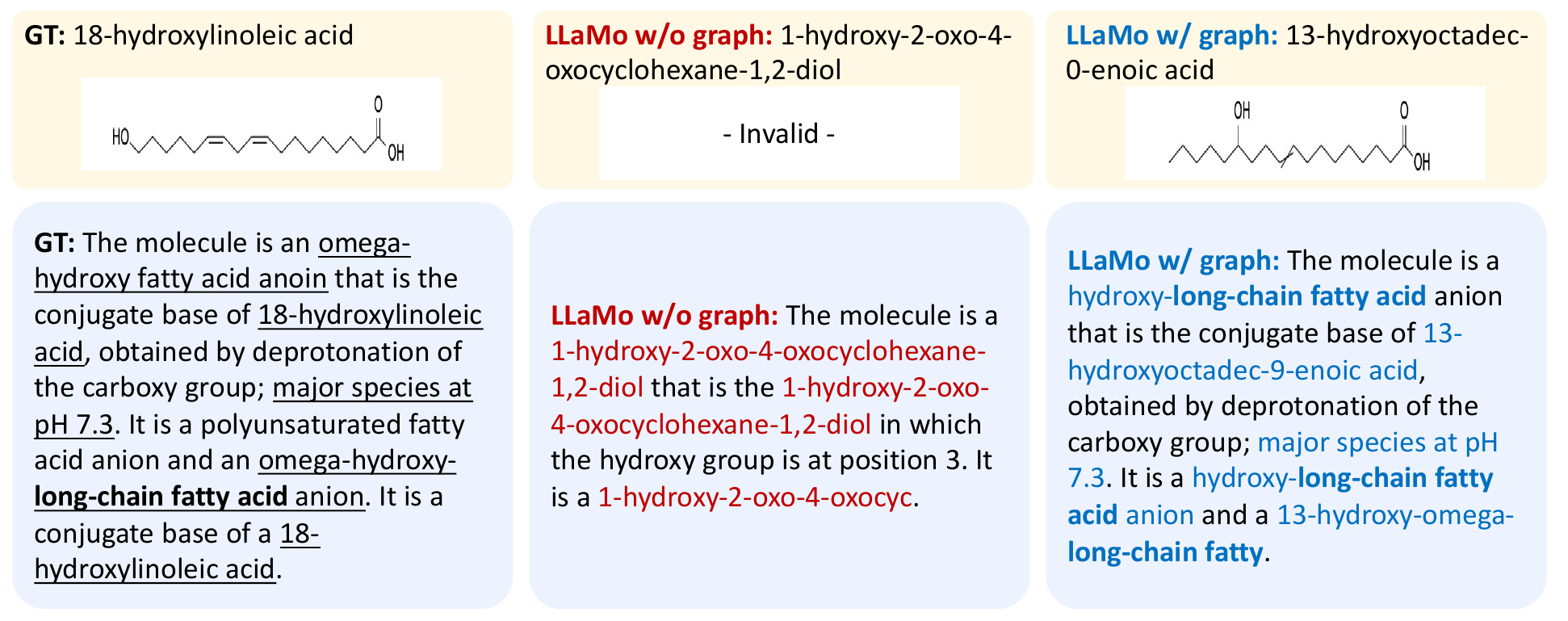}
    \caption{An example of molecular description generation results of LLaMo~w/o graph and LLaMo~w/ graph given the molecule~(‘‘C(CCC/C=C$\backslash$C/C=C$\backslash$CCCCCO)CCCC(=O)[O-1]’’). 
    In the top box, the molecular graphs of IUPAC and functional groups in the descriptions are depicted.
    } 
    \label{fig:qual}
\end{figure}
\begin{figure}[t!]
    \centering
    \includegraphics[width=1.0\textwidth]{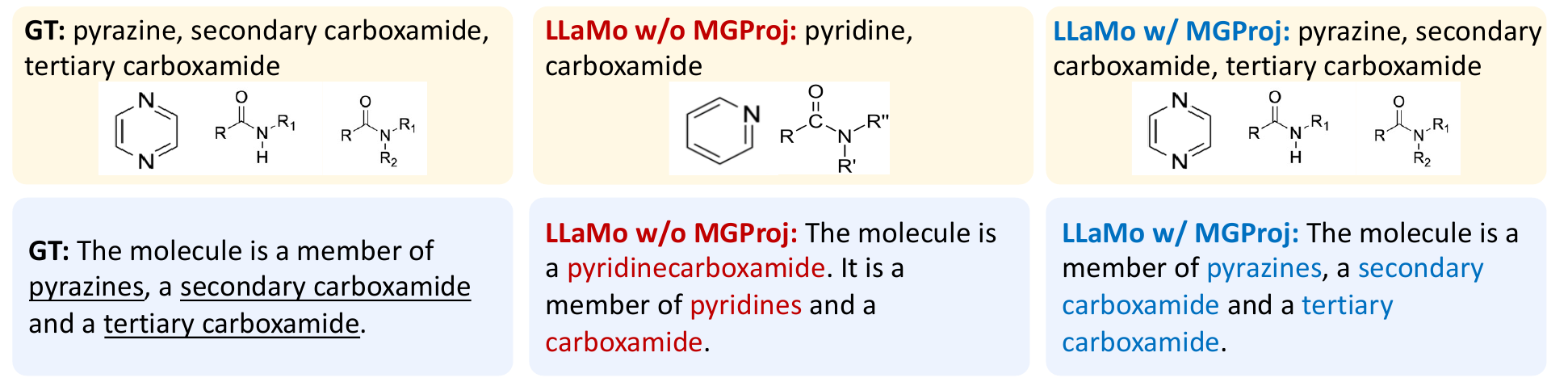}
    \caption{An example of molecular description generation results of LLaMo w/o MGProj and LLaMo w/ MGProj given the molecule 
    (‘‘C[C@@H1]1CN(C(=O)C2=C(C(=CC=C2)NC(=O)C3=NC=CN= C3)O[C@@H1]1CNC)[C@H1](C)CO’’).
    In the top box, the molecular graphs of IUPAC and functional groups in the descriptions are depicted.
    } 
    \label{fig:qual2}
\end{figure}

\noindent\textbf{Qualitative analysis.}
Figure~\ref{fig:qual} shows a GT description and the molecular descriptions generated by the model with and without the molecular graph~(SMILES representation only).
As shown in the figure, LLaMo with a graph denoted as \textbf{LLaMo w/ graph} generates a better molecular description compared to LLaMo without a graph~\textbf{(LLaMo w/o graph)}.
The GT description explains the molecule with `omega-hydroxy-long-chain fatty acid anion'.
Since LLaMo w/o graph does not have any graph structural information, it fails to generate a description with an invalid IUPAC name~(`1-hydroxy-2-oxo-4-oxocyclohexane-1,2-diol'), while LLaMo w/ graph generates a more related description with `hydroxy-long-chain fatty acid anion'.
In addition, we know that LLaMo w/ graph accurately predicts the long-chain structure of the molecule.

We also perform another qualitative analysis by comparing molecular descriptions generated from the model with and without our Multi-level Graph Projector~(MGProj) denoted by \textbf{LLaMo w/ MG Proj} and \textbf{LLaMo w/o MGProj} in Figure~\ref{fig:qual2}.
The figure shows that the multi-level graph projector plays a crucial role in capturing the details of the molecule.
Compared to \textbf{LLaMo w/o MGProj} generating `pyridine', the model with MGProj generates accurate molecular description including `pyrazine' same as GT description.
This demonstrates that the multi-level graph projector is effective in molecule understanding and generation by preserving the molecular graph structural information. 

\section{Conclusion}

We propose LLaMo: Large Language Model-based Molecular graph assistant, an end-to-end trained large molecular graph-language model, to perform various molecule-related tasks with a single model.
For the projector, we newly introduce a multi-level graph projector, which addresses the over-smoothing problem of the graph encoder and captures multi-hop graph information.
We also present machine-generated instruction-following data in the form of multi-turn conversations to improve the instruction-following capabilities of the large language model.

\section*{Acknowledgement}
This work was partly supported by ICT Creative Consilience Program through the Institute of Information \& Communications Technology Planning \& Evaluation (IITP)~(IITP-2024-RS-2020-II201819, \textcolor{black}{10\%}) and the National Research Foundation of Korea (NRF)~(NRF-2023R1A2C2005373, \textcolor{black}{45\%}) grant funded by the Korea government (MSIT), and the Virtual Engineering Platform Project (Grant No. P0022336, \textcolor{black}{45\%}), funded by the Ministry of Trade, Industry \& Energy (MoTIE, South Korea).
We appreciate Dr. Jaesung Kwak for valuable comments and discussions.

{
\small
\bibliography{main}

\begin{thebibliography}{10}

\bibitem{zeng2022deep}
Zheni Zeng, Yuan Yao, Zhiyuan Liu, and Maosong Sun.
\newblock A deep-learning system bridging molecule structure and biomedical text with comprehension comparable to human professionals.
\newblock {\em Nat. Commun.}, 13(1):862, 2022.

\bibitem{taylor2022galactica}
Ross Taylor, Marcin Kardas, Guillem Cucurull, Thomas Scialom, Anthony Hartshorn, Elvis Saravia, Andrew Poulton, Viktor Kerkez, and Robert Stojnic.
\newblock Galactica: A large language model for science.
\newblock {\em arXiv:2211.09085}, 2022.

\bibitem{liu2023molca}
Zhiyuan Liu, Sihang Li, Yanchen Luo, Hao Fei, Yixin Cao, Kenji Kawaguchi, Xiang Wang, and Tat-Seng Chua.
\newblock Molca: Molecular graph-language modeling with cross-modal projector and uni-modal adapter.
\newblock In {\em EMNLP}, 2023.

\bibitem{su2022molecular}
Bing Su, Dazhao Du, Zhao Yang, Yujie Zhou, Jiangmeng Li, Anyi Rao, Hao Sun, Zhiwu Lu, and Ji-Rong Wen.
\newblock A molecular multimodal foundation model associating molecule graphs with natural language.
\newblock {\em arXiv:2209.05481}, 2022.

\bibitem{duvenaud2015convolutional}
David~K Duvenaud, Dougal Maclaurin, Jorge Iparraguirre, Rafael Bombarell, Timothy Hirzel, Al{\'a}n Aspuru-Guzik, and Ryan~P Adams.
\newblock Convolutional networks on graphs for learning molecular fingerprints.
\newblock In {\em NeurIPS}, 2015.

\bibitem{de2018molgan}
Nicola De~Cao and Thomas Kipf.
\newblock Molgan: An implicit generative model for small molecular graphs.
\newblock In {\em ICML workshop}, 2018.

\bibitem{atz2021geometric}
Kenneth Atz, Francesca Grisoni, and Gisbert Schneider.
\newblock Geometric deep learning on molecular representations.
\newblock {\em Nat. Mach. Intell.}, 3(12):1023--1032, 2021.

\bibitem{edwards2021text2mol}
Carl Edwards, ChengXiang Zhai, and Heng Ji.
\newblock Text2mol: Cross-modal molecule retrieval with natural language queries.
\newblock In {\em EMNLP}, 2021.

\bibitem{touvron2023llama}
Hugo Touvron, Louis Martin, Kevin Stone, Peter Albert, Amjad Almahairi, Yasmine Babaei, Nikolay Bashlykov, Soumya Batra, Prajjwal Bhargava, Shruti Bhosale, et~al.
\newblock Llama 2: Open foundation and fine-tuned chat models.
\newblock {\em arXiv:2307.09288}, 2023.

\bibitem{achiam2023gpt}
Josh Achiam, Steven Adler, Sandhini Agarwal, Lama Ahmad, Ilge Akkaya, Florencia~Leoni Aleman, Diogo Almeida, Janko Altenschmidt, Sam Altman, Shyamal Anadkat, et~al.
\newblock Gpt-4 technical report.
\newblock {\em arXiv:2303.08774}, 2023.

\bibitem{team2023gemini}
Gemini Team, Rohan Anil, Sebastian Borgeaud, Yonghui Wu, Jean-Baptiste Alayrac, Jiahui Yu, Radu Soricut, Johan Schalkwyk, Andrew~M Dai, Anja Hauth, et~al.
\newblock Gemini: a family of highly capable multimodal models.
\newblock {\em arXiv:2312.11805}, 2023.

\bibitem{wang2023self}
Yizhong Wang, Yeganeh Kordi, Swaroop Mishra, Alisa Liu, Noah~A Smith, Daniel Khashabi, and Hannaneh Hajishirzi.
\newblock Self-instruct: Aligning language models with self-generated instructions.
\newblock In {\em ACL}, 2023.

\bibitem{chung2024scaling}
Hyung~Won Chung, Le~Hou, Shayne Longpre, Barret Zoph, Yi~Tay, William Fedus, Yunxuan Li, Xuezhi Wang, Mostafa Dehghani, Siddhartha Brahma, et~al.
\newblock Scaling instruction-finetuned language models.
\newblock {\em JMLR}, 25(70):1--53, 2024.

\bibitem{alpaca}
Rohan Taori, Ishaan Gulrajani, Tianyi Zhang, Yann Dubois, Xuechen Li, Carlos Guestrin, Percy Liang, and Tatsunori~B. Hashimoto.
\newblock Stanford alpaca: An instruction-following llama model, 2023.

\bibitem{gao2023llama}
Peng Gao, Jiaming Han, Renrui Zhang, Ziyi Lin, Shijie Geng, Aojun Zhou, Wei Zhang, Pan Lu, Conghui He, Xiangyu Yue, et~al.
\newblock Llama-adapter v2: Parameter-efficient visual instruction model.
\newblock {\em arXiv:2304.15010}, 2023.

\bibitem{zhang2023llama}
Renrui Zhang, Jiaming Han, Chris Liu, Peng Gao, Aojun Zhou, Xiangfei Hu, Shilin Yan, Pan Lu, Hongsheng Li, and Yu~Qiao.
\newblock Llama-adapter: Efficient fine-tuning of language models with zero-init attention.
\newblock In {\em ICLR}, 2024.

\bibitem{dai2023instructblip}
Wenliang Dai, Junnan Li, Dongxu Li, Anthony Meng~Huat Tiong, Junqi Zhao, Weisheng Wang, Boyang Li, Pascale~N Fung, and Steven Hoi.
\newblock Instructblip: Towards general-purpose vision-language models with instruction tuning.
\newblock In {\em NeurIPS}, 2023.

\bibitem{zhu2024minigpt}
Deyao Zhu, Jun Chen, Xiaoqian Shen, Xiang Li, and Mohamed Elhoseiny.
\newblock Minigpt-4: Enhancing vision-language understanding with advanced large language models.
\newblock In {\em ICLR}, 2024.

\bibitem{kipf2016semi}
Thomas~N Kipf and Max Welling.
\newblock Semi-supervised classification with graph convolutional networks.
\newblock In {\em ICLR}, 2017.

\bibitem{velickovic2017graph}
Petar Velickovic, Guillem Cucurull, Arantxa Casanova, Adriana Romero, Pietro Lio, Yoshua Bengio, et~al.
\newblock Graph attention networks.
\newblock In {\em ICLR}, 2018.

\bibitem{xu2019powerful}
Keyulu Xu, Weihua Hu, Jure Leskovec, and Stefanie Jegelka.
\newblock How powerful are graph neural networks?
\newblock In {\em ICLR}, 2019.

\bibitem{park2022deformable}
Jinyoung Park, Sungdong Yoo, Jihwan Park, and Hyunwoo~J Kim.
\newblock Deformable graph convolutional networks.
\newblock In {\em AAAI}, 2022.

\bibitem{ji2022relmole}
Zewei Ji, Runhan Shi, Jiarui Lu, Fang Li, and Yang Yang.
\newblock Relmole: molecular representation learning based on two-level graph similarities.
\newblock {\em J. Chem. Inf. Model.}, 62(22):5361--5372, 2022.

\bibitem{xia2022mole}
Jun Xia, Chengshuai Zhao, Bozhen Hu, Zhangyang Gao, Cheng Tan, Yue Liu, Siyuan Li, and Stan~Z Li.
\newblock Mole-bert: Rethinking pre-training graph neural networks for molecules.
\newblock In {\em ICLR}, 2023.

\bibitem{zhang2021motif}
Zaixi Zhang, Qi~Liu, Hao Wang, Chengqiang Lu, and Chee-Kong Lee.
\newblock Motif-based graph self-supervised learning for molecular property prediction.
\newblock In {\em NeurIPS}, 2021.

\bibitem{zang2023hierarchical}
Xuan Zang, Xianbing Zhao, and Buzhou Tang.
\newblock Hierarchical molecular graph self-supervised learning for property prediction.
\newblock {\em Commun. Chem.}, 6(1):34, 2023.

\bibitem{you2020graph}
Yuning You, Tianlong Chen, Yongduo Sui, Ting Chen, Zhangyang Wang, and Yang Shen.
\newblock Graph contrastive learning with augmentations.
\newblock In {\em NeurIPS}, 2020.

\bibitem{wang2022molecular}
Yuyang Wang, Jianren Wang, Zhonglin Cao, and Amir Barati~Farimani.
\newblock Molecular contrastive learning of representations via graph neural networks.
\newblock {\em Nat. Mach. Intell.}, 4(3):279--287, 2022.

\bibitem{edwards2022translation}
Carl Edwards, Tuan Lai, Kevin Ros, Garrett Honke, Kyunghyun Cho, and Heng Ji.
\newblock Translation between molecules and natural language.
\newblock In {\em EMNLP}, 2022.

\bibitem{zhao2023gimlet}
Haiteng Zhao, Shengchao Liu, Ma~Chang, Hannan Xu, Jie Fu, Zhihong Deng, Lingpeng Kong, and Qi~Liu.
\newblock Gimlet: A unified graph-text model for instruction-based molecule zero-shot learning.
\newblock In {\em NeurIPS}, 2023.

\bibitem{li2018deeper}
Qimai Li, Zhichao Han, and Xiao-Ming Wu.
\newblock Deeper insights into graph convolutional networks for semi-supervised learning.
\newblock In {\em AAAI}, 2018.

\bibitem{wei2021finetuned}
Jason Wei, Maarten Bosma, Vincent~Y Zhao, Kelvin Guu, Adams~Wei Yu, Brian Lester, Nan Du, Andrew~M Dai, and Quoc~V Le.
\newblock Finetuned language models are zero-shot learners.
\newblock In {\em ICLR}, 2022.

\bibitem{ouyang2022training}
Long Ouyang, Jeffrey Wu, Xu~Jiang, Diogo Almeida, Carroll Wainwright, Pamela Mishkin, Chong Zhang, Sandhini Agarwal, Katarina Slama, Alex Ray, et~al.
\newblock Training language models to follow instructions with human feedback.
\newblock In {\em NeurIPS}, 2022.

\bibitem{longpre2023flan}
Shayne Longpre, Le~Hou, Tu~Vu, Albert Webson, Hyung~Won Chung, Yi~Tay, Denny Zhou, Quoc~V Le, Barret Zoph, Jason Wei, et~al.
\newblock The flan collection: Designing data and methods for effective instruction tuning.
\newblock In {\em ICML}, 2023.

\bibitem{sanh2021multitask}
Victor Sanh, Albert Webson, Colin Raffel, Stephen~H Bach, Lintang Sutawika, Zaid Alyafeai, Antoine Chaffin, Arnaud Stiegler, Teven~Le Scao, Arun Raja, et~al.
\newblock Multitask prompted training enables zero-shot task generalization.
\newblock In {\em ICLR}, 2021.

\bibitem{xue2023instruction}
Xue Fuzhao, Jain Kabir, Hitesh~Shah Mahir, Zheng Zangwei, and You Yang.
\newblock Instruction in the wild: A user-based instruction dataset.
\newblock https://github.com/XueFuzhao/InstructionWild, 2023.

\bibitem{vicuna}
Wei-Lin Chiang, Zhuohan Li, Zi~Lin, Ying Sheng, Zhanghao Wu, Hao Zhang, Lianmin Zheng, Siyuan Zhuang, Yonghao Zhuang, Joseph~E. Gonzalez, Ion Stoica, and Eric~P. Xing.
\newblock Vicuna: An open-source chatbot impressing gpt-4 with 90\%* chatgpt quality, 2023.

\bibitem{liu2023visual}
Haotian Liu, Chunyuan Li, Qingyang Wu, and Yong~Jae Lee.
\newblock Visual instruction tuning.
\newblock In {\em NeurIPS}, 2023.

\bibitem{maaz2023video}
Muhammad Maaz, Hanoona Rasheed, Salman Khan, and Fahad~Shahbaz Khan.
\newblock Video-chatgpt: Towards detailed video understanding via large vision and language models.
\newblock In {\em ACL}, 2024.

\bibitem{li2023videochat}
KunChang Li, Yinan He, Yi~Wang, Yizhuo Li, Wenhai Wang, Ping Luo, Yali Wang, Limin Wang, and Yu~Qiao.
\newblock Videochat: Chat-centric video understanding.
\newblock {\em arXiv:2305.06355}, 2023.

\bibitem{weininger1988smiles}
David Weininger.
\newblock Smiles, a chemical language and information system. 1. introduction to methodology and encoding rules.
\newblock {\em J. Chem. Inf. Comput.}, 28(1):31--36, 1988.

\bibitem{ye2023mplug}
Qinghao Ye, Haiyang Xu, Guohai Xu, Jiabo Ye, Ming Yan, Yiyang Zhou, Junyang Wang, Anwen Hu, Pengcheng Shi, Yaya Shi, et~al.
\newblock mplug-owl: Modularization empowers large language models with multimodality.
\newblock {\em arXiv:2304.14178}, 2023.

\bibitem{cha2024honeybee}
Junbum Cha, Wooyoung Kang, Jonghwan Mun, and Byungseok Roh.
\newblock Honeybee: Locality-enhanced projector for multimodal llm.
\newblock In {\em CVPR}, 2024.

\bibitem{kim2021pubchem}
Sunghwan Kim, Jie Chen, Tiejun Cheng, Asta Gindulyte, Jia He, Siqian He, Qingliang Li, Benjamin~A Shoemaker, Paul~A Thiessen, Bo~Yu, et~al.
\newblock Pubchem in 2021: new data content and improved web interfaces.
\newblock {\em Nucleic Acids Res.}, 49(D1):D1388--D1395, 2021.

\bibitem{hu2022lora}
Edward~J Hu, Yelong Shen, Phillip Wallis, Zeyuan Allen-Zhu, Yuanzhi Li, Shean Wang, Lu~Wang, and Weizhu Chen.
\newblock Lora: Low-rank adaptation of large language models.
\newblock In {\em ICLR}, 2022.

\bibitem{xu2023lvlm}
Peng Xu, Wenqi Shao, Kaipeng Zhang, Peng Gao, Shuo Liu, Meng Lei, Fanqing Meng, Siyuan Huang, Yu~Qiao, and Ping Luo.
\newblock Lvlm-ehub: A comprehensive evaluation benchmark for large vision-language models.
\newblock {\em arXiv:2306.09265}, 2023.

\bibitem{li2023textbind}
Huayang Li, Siheng Li, Deng Cai, Longyue Wang, Lemao Liu, Taro Watanabe, Yujiu Yang, and Shuming Shi.
\newblock Textbind: Multi-turn interleaved multimodal instruction-following.
\newblock In {\em ACL findings}, 2024.

\bibitem{fang2024mol}
Yin Fang, Xiaozhuan Liang, Ningyu Zhang, Kangwei Liu, Rui Huang, Zhuo Chen, Xiaohui Fan, and Huajun Chen.
\newblock Mol-instructions: A large-scale biomolecular instruction dataset for large language models.
\newblock In {\em ICLR}, 2024.

\bibitem{christofidellis2023unifying}
Dimitrios Christofidellis, Giorgio Giannone, Jannis Born, Ole Winther, Teodoro Laino, and Matteo Manica.
\newblock Unifying molecular and textual representations via multi-task language modelling.
\newblock In {\em ICML}, 2023.

\bibitem{bai2023qwen}
Jinze Bai, Shuai Bai, Shusheng Yang, Shijie Wang, Sinan Tan, Peng Wang, Junyang Lin, Chang Zhou, and Jingren Zhou.
\newblock Qwen-vl: A frontier large vision-language model with versatile abilities.
\newblock {\em arXiv:2308.12966}, 2023.

\bibitem{xu2023baize}
Canwen Xu, Daya Guo, Nan Duan, and Julian McAuley.
\newblock Baize: An open-source chat model with parameter-efficient tuning on self-chat data.
\newblock In {\em EMNLP}, 2023.

\bibitem{glm2024chatglm}
Team GLM, Aohan Zeng, Bin Xu, Bowen Wang, Chenhui Zhang, Da~Yin, Diego Rojas, Guanyu Feng, Hanlin Zhao, Hanyu Lai, et~al.
\newblock Chatglm: A family of large language models from glm-130b to glm-4 all tools.
\newblock {\em arXiv:2406.12793}, 2024.

\bibitem{touvron2023llama1}
Hugo Touvron, Thibaut Lavril, Gautier Izacard, Xavier Martinet, Marie-Anne Lachaux, Timoth{\'e}e Lacroix, Baptiste Rozi{\`e}re, Naman Goyal, Eric Hambro, Faisal Azhar, et~al.
\newblock Llama: Open and efficient foundation language models.
\newblock {\em arXiv:2302.13971}, 2023.

\bibitem{cao2023instructmol}
He~Cao, Zijing Liu, Xingyu Lu, Yuan Yao, and Yu~Li.
\newblock Instructmol: Multi-modal integration for building a versatile and reliable molecular assistant in drug discovery.
\newblock {\em arXiv:2311.16208}, 2023.

\bibitem{paszke2019pytorch}
Adam Paszke, Sam Gross, Francisco Massa, Adam Lerer, James Bradbury, Gregory Chanan, Trevor Killeen, Zeming Lin, Natalia Gimelshein, Luca Antiga, et~al.
\newblock Pytorch: An imperative style, high-performance deep learning library.
\newblock In {\em NeurIPS}, 2019.

\bibitem{fey2019fast}
Matthias Fey and Jan~Eric Lenssen.
\newblock Fast graph representation learning with pytorch geometric.
\newblock In {\em ICLR workshop}, 2019.

\bibitem{wolf2019huggingface}
Thomas Wolf, Lysandre Debut, Victor Sanh, Julien Chaumond, Clement Delangue, Anthony Moi, Pierric Cistac, Tim Rault, R{\'e}mi Louf, Morgan Funtowicz, et~al.
\newblock Huggingface's transformers: State-of-the-art natural language processing.
\newblock In {\em EMNLP demo}, 2020.

\bibitem{peft}
Sourab Mangrulkar, Sylvain Gugger, Lysandre Debut, Younes Belkada, Sayak Paul, and Benjamin Bossan.
\newblock Peft: State-of-the-art parameter-efficient fine-tuning methods, 2022.

\bibitem{hu2023opendelta}
Shengding Hu, Ning Ding, Weilin Zhao, Xingtai Lv, Zhen Zhang, Zhiyuan Liu, and Maosong Sun.
\newblock Opendelta: A plug-and-play library for parameter-efficient adaptation of pre-trained models.
\newblock In {\em ACL}, 2023.

\bibitem{liu2023multi}
Shengchao Liu, Weili Nie, Chengpeng Wang, Jiarui Lu, Zhuoran Qiao, Ling Liu, Jian Tang, Chaowei Xiao, and Animashree Anandkumar.
\newblock Multi-modal molecule structure--text model for text-based retrieval and editing.
\newblock {\em Nat. Mach. Intell.}, 5(12):1447--1457, 2023.

\bibitem{radford2021learning}
Alec Radford, Jong~Wook Kim, Chris Hallacy, Aditya Ramesh, Gabriel Goh, Sandhini Agarwal, Girish Sastry, Amanda Askell, Pamela Mishkin, Jack Clark, et~al.
\newblock Learning transferable visual models from natural language supervision.
\newblock In {\em ICML}, 2021.

\bibitem{hu2020open}
Weihua Hu, Matthias Fey, Marinka Zitnik, Yuxiao Dong, Hongyu Ren, Bowen Liu, Michele Catasta, and Jure Leskovec.
\newblock Open graph benchmark: Datasets for machine learning on graphs.
\newblock In {\em NeurIPS}, 2020.

\bibitem{loshchilov2017decoupled}
Ilya Loshchilov and Frank Hutter.
\newblock Decoupled weight decay regularization.
\newblock In {\em ICLR}, 2019.

\bibitem{wu2018moleculenet}
Zhenqin Wu, Bharath Ramsundar, Evan~N Feinberg, Joseph Gomes, Caleb Geniesse, Aneesh~S Pappu, Karl Leswing, and Vijay Pande.
\newblock Moleculenet: a benchmark for molecular machine learning.
\newblock {\em Chem. Sci.}, 9(2):513--530, 2018.

\bibitem{papineni2002bleu}
Kishore Papineni, Salim Roukos, Todd Ward, and Wei-Jing Zhu.
\newblock Bleu: a method for automatic evaluation of machine translation.
\newblock In {\em ACL}, 2002.

\bibitem{banerjee2005meteor}
Satanjeev Banerjee and Alon Lavie.
\newblock Meteor: An automatic metric for mt evaluation with improved correlation with human judgments.
\newblock In {\em ACL workshop}, 2005.

\bibitem{tanimoto1958elementary}
Taffee~T Tanimoto.
\newblock {\em Elementary mathematical theory of classification and prediction}.
\newblock International Business Machines Corp., 1958.

\bibitem{morgan1965generation}
Harry~L Morgan.
\newblock The generation of a unique machine description for chemical structures-a technique developed at chemical abstracts service.
\newblock {\em J. Chem. Doc.}, 5(2):107--113, 1965.

\bibitem{liu2023molxpt}
Zequn Liu, Wei Zhang, Yingce Xia, Lijun Wu, Shufang Xie, Tao Qin, Ming Zhang, and Tie-Yan Liu.
\newblock Molxpt: Wrapping molecules with text for generative pre-training.
\newblock In {\em ACL}, 2023.

\bibitem{luo2023molfm}
Yizhen Luo, Kai Yang, Massimo Hong, Xingyi Liu, and Zaiqing Nie.
\newblock Molfm: A multimodal molecular foundation model.
\newblock {\em arXiv:2307.09484}, 2023.

\bibitem{liu2024git}
Pengfei Liu, Yiming Ren, Jun Tao, and Zhixiang Ren.
\newblock Git-mol: A multi-modal large language model for molecular science with graph, image, and text.
\newblock {\em Comput. Biol. Med.}, 171:108073, 2024.

\bibitem{hastings2016chebi}
Janna Hastings, Gareth Owen, Adriano Dekker, Marcus Ennis, Namrata Kale, Venkatesh Muthukrishnan, Steve Turner, Neil Swainston, Pedro Mendes, and Christoph Steinbeck.
\newblock Chebi in 2016: Improved services and an expanding collection of metabolites.
\newblock {\em Nucleic Acids Res.}, 44(D1):D1214--D1219, 2016.

\bibitem{favre2013nomenclature}
Henri~A Favre and Warren~H Powell.
\newblock {\em Nomenclature of organic chemistry: IUPAC recommendations and preferred names 2013}.
\newblock Royal Society of Chemistry, 2013.

\bibitem{wei2010novel}
Jin-Mao Wei, Xiao-Jie Yuan, Qing-Hua Hu, and Shu-Qin Wang.
\newblock A novel measure for evaluating classifiers.
\newblock {\em Expert Syst. Appl.}, 37(5):3799--3809, 2010.

\bibitem{lu2022unified}
Jieyu Lu and Yingkai Zhang.
\newblock Unified deep learning model for multitask reaction predictions with explanation.
\newblock {\em J. Chem. Inf. Model.}, 62(6):1376--1387, 2022.

\end{thebibliography}
\bibliographystyle{unsrt}
}
\clearpage
\appendix
The appendix is organized into the following sections.
\begin{outline}
    \1 Appendix \ref{app_sec:limitations}: Limitations
    \1 Appendix \ref{app_sec:broader}: Broader Impacts
    \1 Appendix \ref{app_sec:gnn}: Explanation on Graph Neural Networks
    \1 Appendix \ref{app_sec:add}: Additional Experimental Results
    \1 Appendix \ref{app_sec:detailed}: Detailed Experimental Settings
    \2 Appendix \ref{sup:implementation_details}: Implementation Details
    \2 Appendix \ref{sup:metrics}: Metrics
    \1 Appendix \ref{app_sec:baselines}: Baselines 
    \1 Appendix \ref{app_sec:datasets}: Benchmarks
    \1 Appendix \ref{app_sec:prompts}: Prompts
    \1 Appendix \ref{app_sec:oversmoothing}: Over-smoothing Problems
    \1 Appendix \ref{app_sec:qual}: More Qualitative Samples
\end{outline}
\section{Limitations}
\label{app_sec:limitations}
Our LLaMo is built upon an LLM, \textit{e.g.}, LLaMA and Galactica, and is fine-tuned on molecule benchmark datasets by leveraging its pretrained knowledge.
Given that LLMs are pretrained using extensive web-crawled corpora, it is uncertain whether the data used for LLMs' pretraining and the test samples in molecule benchmark datasets are mutually exclusive.
This results in implicit data leakage when fine-tuning and evaluating LLaMo on molecule benchmark datasets.
Furthermore, LLMs inherently require large memory and computational costs and cause hallucination problems where the model generates incorrect but plausible text.
Our LLaMo may inherit these LLMs' problems due to LLMs' powerful pre-trained knowledge.
\section{Broader Impacts}
\label{app_sec:broader}
We proposed the first molecular graph-based general-purpose model, LLaMo, which is widely applicable to various molecule tasks such as molecule captioning, property prediction, and IUPAC naming.
Our LLaMo itself does not have negative societal impacts.
However, as discussed above, since our model is built upon an LLM, the model sometimes generates biased output concerning race, religion, culture, and gender, resulting in the misusage of our model.
Also, training LLMs requires massive amounts of CO2 emission promoting global warming.

\section{Explanation on Graph Neural Networks}
\label{app_sec:gnn}
Let $\mathcal{G}=\left(\mathcal{V}, \mathcal{E}, \mathbf{X} \right)$ denote the input graph, where $\mathcal{V}, \mathcal{E}$ are a set of nodes and edges, respectively, and $\mathbf{X}$ indicates a set of input node features. 
The input feature of node $v \in \mathcal{V}$ is defined as $\mathbf{x}_v$ and the edge between node $u$ and $v$ is represented with $\left(u, v \right)\in \mathcal{V} \times \mathcal{V}$.
The neighbor set of node $v$ on the input graph is denoted by $\mathcal{N}_v = \left\{u | \left(u, v \right) \in \mathcal{E} \right\}$.
Given the graph, graph neural networks iteratively update node representation $\mathbf{z}_v^{(l)} \in \mathbb{R}^{d^{(l)}}$ via the following message-passing framework:
\begin{equation}
    \mathbf{z}_v^{(l)} = \text{UPDATE}^{(l)} \left(\mathbf{z}_v^{(l-1)},  \ \text{AGGREGATE}^{(l)}\left(\left\{\mathbf{z}_u^{(l-1)}: u \in \mathcal{N}_v \right\}\right) \right), \quad l=1, \dots L
    \label{eq:gnn}
\end{equation}
where $\mathbf{z}_v^{(0)} = \mathbf{x}_v$, in that the node representation of $0$-th layer is input node features.
$\text{AGGREGATE}\left(\cdot \right)$ function aggregates the representations of the neighbor set with a particular function.
$\text{UPDATE}\left(\cdot \right)$ function is designed to update a node reprsentation $\mathbf{z}_v^{(l-1)}$ with the aggregated information produced by $\text{AGGREGATE}\left(\cdot \right)$.
With the message-passing, $L$-layer GNN provides node representations $\mathbf{z}_v^{(L)}$ that express an $L$-hop egograph given the node $v$ as a center node.
In this paper, we use a pre-trained GIN~\cite{xu2019powerful} with 5 layers for the graph encoder, which has widely been applied to molecule graph understanding tasks~\cite{su2022molecular}. 
Formally, GIN updates node representations $\mathbf{z}_v^{(l)}$ with the following equation:
\begin{equation}
    \mathbf{z}_v^{(l)} = \text{MLP}^{(l)} \left( \left(1+ \epsilon^{(l)} \right)\cdot \mathbf{z}_v^{(l-1)} + \sum_{u \in \mathcal{N}_v} \mathbf{z}_u^{(l-1)} \right),
\end{equation}
where $\epsilon^{(l+1)}$ is a learnable parameter or a fixed scalar.
The aggregate function $\text{AGGREGATE}\left(\cdot\right)$ is the sum operation and the MLP function is used for the combine function $\text{UPDATE}\left(\cdot\right)$.

\section{Additional Experimental Results}
\label{app_sec:add}
\begin{table*}[t]
\centering
\caption{Performance on chemical reaction tasks, including forward reaction prediction and retrosynthesis. $*$ denotes the model fine-tuned with task-specific instruction data.}
\label{tab:chemical_reaction}
\setlength{\tabcolsep}{4.3mm}{
\resizebox{\textwidth}{!}{
\begin{tabular}{lccccccc}
\toprule
  {Model}
&  {Exact}$\uparrow$  &   {BLEU}$\uparrow$  &   {Levenshtein}$\downarrow$  &   {RDK FTS}$\uparrow$  &   {MACCS FTS}$\uparrow$ &   {Morgan FTS}$\uparrow$ &   {Validity}$\uparrow$ \\
\midrule
\midrule[1.1pt]
\rowcolor[RGB]{234, 238, 234}
\multicolumn{8}{l}{\textit{Forward Reaction Prediction}} \\
Alpaca$^\dagger$~\cite{alpaca} & 0.000 & 0.065 & 41.989 & 0.004 & 0.024 & 0.008 & 0.138 \\
Baize$^\dagger$~\cite{xu2023baize} & 0.000 & 0.044 & 41.500 & 0.004 & 0.025 & 0.009 & 0.097 \\
ChatGLM$^\dagger$~\cite{glm2024chatglm} & 0.000 & 0.183 & 40.008 & 0.050 & 0.100 & 0.044 & 0.108 \\
LLaMA$^\dagger$~\cite{touvron2023llama1} & 0.000 & 0.020 & 42.002 & 0.001 & 0.002 & 0.001 & 0.039 \\
Vicuna$^\dagger$~\cite{vicuna} & 0.000 & 0.057 & 41.690 & 0.007 & 0.016 & 0.006 & 0.059 \\
\midrule
LLaMA$^*$~\cite{touvron2023llama1} &0.012	&0.804	&29.947	&0.499	&0.649	&0.407	&\textbf{1.000} \\
Mol-Instruction~\cite{fang2024mol} & 0.045 & 0.654 & 27.262 & 0.313 & 0.509 & 0.262 & \textbf{1.000} \\
{InstructMol-G}~\cite{cao2023instructmol}        & 0.153 &0.906 &20.155	&0.519	&0.717	&0.457	&\textbf{1.000} \\
{InstructMol-GS}~\cite{cao2023instructmol} & {0.536} &\textbf{0.967}	&{10.851}	&{0.776}	&{0.878}	&{0.741}	&\textbf{1.000}\\
\midrule
\textbf{LLaMo {(Ours)}} &\textbf{0.584} & 0.894 & \textbf{6.162} & \textbf{0.857} & \textbf{0.918} & \textbf{0.841} & 0.938\\

\midrule[1.1pt]
\rowcolor[RGB]{234, 238, 234}
\multicolumn{8}{l}{\textit{Retrosynthesis}} \\
Alpaca$^\dagger$~\cite{alpaca} & 0.000 & 0.063 & 46.915 & 0.005 & 0.023 & 0.007 & 0.160 \\
Baize$^\dagger$~\cite{xu2023baize} & 0.000 & 0.095 & 44.714 & 0.025 & 0.050 & 0.023 & 0.112 \\
ChatGLM$^\dagger$~\cite{glm2024chatglm} & 0.000 & 0.117 & 48.365 & 0.056 & 0.075 & 0.043 & 0.046 \\
LLaMA$^\dagger$~\cite{touvron2023llama1} & 0.000 & 0.036 & 46.844 & 0.018 & 0.029 & 0.017 & 0.010 \\
Vicuna$^\dagger$~\cite{vicuna} & 0.000 & 0.057 & 46.877 & 0.025 & 0.030 & 0.021 & 0.017 \\
\midrule
LLaMA$^*$~\cite{touvron2023llama1} &0.000	&0.283	&53.510	&0.136	&0.294	&0.106	&\textbf{1.000} \\
Mol-Instruction~\cite{fang2024mol} & 0.009 & 0.705 & 31.227 & 0.283 & 0.487 & 0.230 & \textbf{1.000} \\
InstructMol-G~\cite{cao2023instructmol}      & 0.114	&0.586	&21.271	&0.422	&0.523	&0.285	&\textbf{1.000}\\
InstructMol-GS~\cite{cao2023instructmol} & \textbf{0.407}	&\textbf{0.941}	&{13.967}	&{0.753}	&{0.852}	&{0.714}	&\textbf{1.000}\\
\midrule
\textbf{LLaMo (Ours)} & 0.341 & 0.830 & \textbf{12.263} & \textbf{0.793} & \textbf{0.868} & \textbf{0.750} & 0.954 \\
\bottomrule
\end{tabular}
}}
\end{table*}

We also conduct experiments to validate the effectiveness of our LLaMo on chemical reaction prediction, such as forward reaction prediction and retrosynthesis tasks in Table~\ref{tab:chemical_reaction}.
The table demonstrates that our LLaMo still performs well on the chemical reaction tasks.
LLaMo achieves the best performance on Levenshtein, RDK FTS, MACCS FTS, and Morgan FTS metrics across diverse baselines in all tasks, which indicates that LLaMo successfully comprehends molecular graph structure.
We conjecture that our multi-level graph projector helps the large language model understand the molecular graph structure by representing multi-hop structural information to the molecular graph tokens.
\section{Detailed Experimental Settings}
\label{app_sec:detailed}

\subsection{Implementation Details}
\label{sup:implementation_details}
Our code is implemented based on PyTorch~\cite{paszke2019pytorch} library. Also, we adopt PyTorch Geometric (PyG)~\cite{fey2019fast}, and Huggingface transformers~\cite{wolf2019huggingface} to utilize the graph architectures and Large Language Models (LLMs).
PEFT~\cite{peft} and OpenDelta~\cite{hu2023opendelta} libraries are used for parameter-efficient fine-tuning of LLMs, \textit{i.e.}, LoRA.
We use LLaMA 2 chat 7B model~\cite{touvron2023llama} and Galactica 1.3B~\cite{taylor2022galactica} as our base language model.
We leverage GIN~\cite{xu2019powerful} with five layers initialized based on the MoleculeSTM graph encoder~\cite{liu2023multi}, which is pre-trained with text-graph contrastive learning~\cite{radford2021learning}.
We use LoRA to train the large language model in stage 2.
We use OGB~\cite{hu2020open}, a smiles2graph function, to convert SMILES representations to 2D graphs.
Our experiments are run on 4 $\times$ A6000 GPUs or 4 $\times$ V100 GPUs and 2 $\times$ A6000 GPUs for LLaMA2 and Galactica, respectively.
In stage 1, the AdamW~\cite{loshchilov2017decoupled} optimizer is adapted with an initial learning rate of 1e-4 (minimum learning rate is 1e-5 and warmup learning rate is 1e-6).
The warmup step is 1,000 and the cosine scheduler is applied.
In stage 2, the initial learning rate is set to 5e-5 (minimum learning rate is 5e-6 and warmup learning rate is 5e-7).

To evaluate the efficacy of the proposed method, we fine-tune baseline models and evaluate them for three tasks such as \textbf{1)} molecule description generation, \textbf{2)} IUPAC name prediction, and \textbf{3)} property question answering.
We conducted experiments under two major settings: generalist and specialist models.
In the generalist setting, one model handles all three tasks, whereas in the specialist setting, we train a model for each downstream task.
For generalist experiments, we use datasets derived from PubChem and QM9. 
To be specific, for `molecule description generation', and `property prediction', 
we use the datasets derived from PubChem and QM9 of MoleculeNet~\cite{wu2018moleculenet} as in Mol-Instructions~\cite{fang2024mol}.
For IUPAC name prediction, a dataset derived from \cite{liu2023molca} is used. 
To train the generalist variant of LLaMo, we use a training split of molecular description generation dataset of Mol-Instructions in stage 1.
In stage 2, the model is instruction-tuned with a training split of description generation and property prediction instruction dataset of Mol-Instructions, IUPAC name prediction from \cite{liu2023molca}, and our \textcolor{black}{GPT-generated} instruction-following data.
For the evaluation of molecular description generation and property question answering tasks, we use the test split of Mol-Instructions molecular description generation and property prediction datasets, which are sampled from PubChem~\cite{kim2021pubchem} and QM9 dataset of MoleculeNet~\cite{wu2018moleculenet}, respectively.
For ablations, we use a short training schedule~(epoch 1 pre-training, epoch 1 instruction tuning).
For the final models, we adopt a long training schedule~(epoch 1 pre-training, epoch 3 instruction tuning).

For training the specialist variant of LLaMo, we follow MolCA~\cite{liu2023molca} to train the model with the pretrain split of PubChem324k in the stage 1 phase and fine-tune the model for each specific downstream task in stage 2.
For the inference under specialist experiments, where a model is individually finetuned for a specific downstream task, we use a test split of PubChem324k~\cite{liu2023molca}, ChEBI-20~\cite{edwards2022translation} and IUPAC name prediction dataset from \cite{liu2023molca}.

\subsection{Metrics}
\label{sup:metrics}
We report BLEU~\cite{papineni2002bleu} and METEOR~\cite{banerjee2005meteor} for the molecule description generation and IUPAC name prediction tasks.
MAE is reported for property QA.

\noindent \textbf{BLEU.}
The BLEU metric measures the quality of generated text by comparing $n$-gram sequence between the generated text and the reference text, which can be formulated as:
\begin{equation}
    \text{BLEU} = \text{BP} \times \exp(\frac{1}{N}\sum_{n=1}^N\log p_n),
\end{equation}
where $N$ is the number of $n$-grams and $p_n$ is the precision, \textit{i.e.}, the ratio of the number of $n$-grams in the generated text appearing in the reference text.
The BLEU score also takes into account sequence length with Brevity Penalty (BP) as:
\begin{equation}
    \text{BP} = 
    \begin{cases} 
        1 & \text{if $c > r$} \\ 
        e^{(1 - r / c)} & \text{if $c \leq r$}
    \end{cases},
\end{equation}
where $c$ and $r$ are the lengths of generated and reference texts, respectively.
This encourages the model to avoid generating short sequences.
In our experiments, we use BLEU-4 as the default BLEU metric.

\noindent \textbf{METEOR.}
The METEOR metric is proposed to consider both precision and recall between the generated text and the reference text, which is as follows:
\begin{equation}
    P = \frac{\text{number of matched words}}{\text{number of words in generated text}}, \:\:\: 
    R = \frac{\text{number of matched words}}{\text{number of words in reference text}},
\end{equation}

\begin{equation}
    F = \frac{10PR}{9P + R},
\end{equation}

\begin{equation}
    \text{Penalty} = 0.5 \cdot \left(\frac{\text{number of chunks}}{\text{number of matched words}}\right),
\end{equation}

\begin{equation}
    \text{METEOR} = F \cdot (1 - \text{Penalty}),
\end{equation}
where a chunk is a set of uni-grams which are adjacent in the generated text and in the reference text.
Similar to BLEU, the Penalty is reflected to take into account the length of generated text.
Therefore, the METEOR metric is specialized to measure the morphology, fluency, and adequacy of text rather than sequence order since it does not use $n$-grams.

\noindent \textbf{MAE.}
Mean Absolute Error (MAE) aims to measure the average magnitude of errors between the ground-truth values and predicted values, which is defined as:
\begin{equation}
    \text{MAE} = \frac{1}{N}\sum_{n=1}^N |\hat{y}_n - y_n|.
\end{equation}
where $y_n$ is a ground-truth and $\hat{y}_n$ is a model prediction.
\section{Baselines}
\label{app_sec:baselines}
For the generalist models, we train our LLaMo based on \texttt{Llama-2-7b-chat}~\cite{touvron2023llama} as a backbone language model for a fair comparison with Mol-Instructions~\cite{fang2024mol}.
We compare our LLaMo with (1) LLM-based generalist models including Galactica~\cite{taylor2022galactica}, \texttt{Llama2-7b-chat}~\cite{touvron2023llama}, GPT-3.5, and GPT-4, (2) Molecule instruction-tuned generalist model such as Mol-Instructions~\cite{fang2024mol}.
Since GPT-3.5 and GPT-4 have difficulty in solving the tasks without in-context learning, we additionally measure the performance of GPT-3.5 and GPT-4 with 4-shots in-context learning, which are GPT-3.5 (ICL) and GPT-4 (ICL). The 4-shot exemplars are selected by computing the Tanimoto similarity~\cite{tanimoto1958elementary} using a 2048-bit Morgan Fingerprint~\cite{morgan1965generation} with RDKit\footnote{\url{https://github.com/rdkit/rdkit}. Copyright (c) 2006-2015, Rational Discovery LLC, Greg Landrum, and Julie Penzotti and others. Licensed under BSD 3-Clause license}, choosing the four molecules most similar to the target molecule from the train split of each dataset.
For the specialist models, we train our LLaMo with Galactica 1.3B~\cite{taylor2022galactica} for a fair comparison with MolCA~\cite{liu2023molca}.
We use single-task specialist models as baselines, including MolT5~\cite{edwards2022translation}, MoMu~\cite{su2022molecular}, and MolCA~\cite{liu2023molca}.
\section{Benchmarks}
\label{app_sec:datasets}
\label{sup:datasets}
In this section, we provide a brief introduction to each task and dataset utilized in our research.

\noindent\textbf{Molecular description generation.} Generating the description of a molecule is considered one of the most important tasks in molecular language models. For a given molecule, we aim for the model to generate an accurate and informative description including various chemical properties, functional groups, biological roles, and real-world applications of the molecule. Developing a model capable of generating such descriptions is highly valuable because it has the potential to discover information about molecules that is currently unknown or very expensive to find out, thus serving as a powerful assistant for various tasks in chemistry and biology. Therefore, various works~\cite{edwards2022translation, su2022molecular, liu2023molxpt, luo2023molfm, liu2023molca, liu2024git} have tried to enhance the ability to generate appropriate descriptions of chemical compounds using language models or multi-modal language models.

For testing the performance of molecule description generation of LLaMo and previous models, we utilize the molecular description generation dataset of Mol-Instructions\footnote{Copyright (c) 2023 ZJUNLP. Licensed under CC-BY 4.0 license}~\cite{fang2024mol}, based on PubChem database~\cite{kim2021pubchem} for generalist models, and both PubChem324k~\cite{liu2023molca} and ChEBI-20~\cite{edwards2021text2mol} for specialist models. PubChem324k is constructed by collecting 324k molecules and their associated text information from the PubChem database. ChEBI-20 is the most commonly utilized benchmark in this task, consisting of selected 33,010 pairs of molecules and descriptions from ChEBI~\cite{hastings2016chebi}. Each description of ChEBI-20 contains more than 20 words and includes various and rich information about molecules, such as conjugate base/acid, functional parent, and enantiomer of molecules. We employ a test split of each dataset: 1,000 samples of Mol-Instructions and 2,000 / 3,300 samples of PubChem324k / ChEBI-20. 
Similarly, we filter out samples of which SMILES representation cannot be converted into the molecular graph via RDKit.

\noindent\textbf{IUPAC name prediction.}IUPAC (International Union of Pure and Applied Chemistry) nomenclature~\cite{favre2013nomenclature} provides a systematic naming convention for molecules based on pre-defined rules, eliminating the ambiguity in molecular names. It is the standard method for naming molecular compounds in chemistry, and as such, numerous pieces of chemical literature utilize IUPAC names to represent molecules. Consequently, previous studies~\cite{taylor2022galactica, liu2023molca} have adopted the task of predicting IUPAC names from SMILES representations to evaluate the chemical understanding capability of molecular language models. We also follow this approach in our research. To assess the performance of LLaMo and baseline models, we use the PubChem324k dataset~\cite{liu2023molca} again.
The test split of PubChem324k offers curated high-quality samples of 2,000 molecules, each with corresponding SMILES representations and IUPAC names.
We provide the SMILES representation of each molecule to the model along with a prompt and then compare the model-generated IUPAC names with the ground-truth IUPAC names to evaluate performance.

\noindent\textbf{Property prediction.} In chemical and biological domains, particularly in drug discovery, exploring chemical compounds that satisfy specific chemical properties is crucial. Therefore, the ability to estimate the chemical and physical properties is essential for chemical foundation models. Additionally, leveraging machine learning-based approaches for predicting molecular properties has proven to be much more efficient than traditional approaches in computational chemistry. Inspired by these facts, property prediction with LLMs has recently drawn attention from researchers~\cite{taylor2022galactica, liu2023molxpt, luo2023molfm, liu2024git}. So, we also assess the property QA performance of LLaMo and baselines using the dataset introduced by \cite{fang2024mol}. This dataset is a subset of QM9 dataset from MoleculeNet \cite{wu2018moleculenet}, a widely used benchmark for various chemistry machine learning tasks. The original QM9 dataset contains numerical values for 19 chemical properties per molecule, but~\cite{fang2024mol} focuses on three properties related to molecular orbital energy: HOMO (Highest Occupied Molecular Orbital) energy, LUMO (Lowest Unoccupied Molecular Orbital) energy, and the HOMO-LUMO gap (in Hartree units).~\cite{fang2024mol} also created distinct question-form instructions for each property to make language models understand the task and accurately generate continuous values.
Consequently, using the given instruction for each molecule in the 2,000 test samples and its SMILES representation as input, we have LLaMo and baseline generalist models predict the properties of the given molecules.
Similarly, we filter out samples of which SMILES representation cannot be converted into the molecular graph via RDKit.

\noindent\textbf{Forward reaction prediction.}
Understanding chemical reactions is crucial for advancing various chemical and biological fields, including pharmaceuticals, materials science, and environmental technology. This understanding enables researchers to design efficient synthesis pathways and develop new chemical compounds. Therefore, if a molecular model can effectively comprehend these chemical reactions, it can serve as a powerful assistant in the research and development process. This concept has emerged through the task of forward reaction prediction, and several prior models~\cite{cao2023instructmol, fang2024mol} have conducted experiments to address this task. This task focuses on the proactive determination of potential products resulting from a chemical reaction based on given reactants and reagents. This approach is also significant from the perspectives of efficiency and environmental sustainability, as it reduces the need for experimental trial and error in chemical development and research by using these models to virtually conduct simulated experiments.
We utilize the forward reaction prediction segment of Mol-Instructions~\citep{fang2024mol}, based on the USPTO~\citep{wei2010novel} database, to assess the performance of LLaMo and other models. This dataset contains question-form instructions for predicting reaction products, with reactants and reagents separated by a period (‘.’) as input, and the corresponding product of the reaction as the target output. It also doesn’t specify what the reactants and reagents are to create a task that more closely resembles real-world scenarios. 
Finally, using the test samples from this dataset, we provide the SMILES representations of the reactants and reagents to the model and their molecular graphs with a prompt, aiming to predict the product molecule in SMILES format as well.

\noindent\textbf{Retrosynthesis.}
Retrosynthesis shares a similar context with forward reaction prediction but approaches chemical reactions from a different perspective. The retrosynthetic analysis begins with the target compound and works backward to identify potential reactant molecules for its synthesis. This reverse approach is as valuable as forward reaction synthesis because it aids researchers in discovering effective and efficient synthetic methodologies for generating target molecules. This is particularly important in various chemical applications, such as drug discovery, where identifying chemically valid and economical processes for synthesizing target drug molecules is essential.
As a result, several previous studies~\citep{cao2023instructmol, fang2024mol} have adopted this task to analyze the capability to deeply understand chemical knowledge and take advantage of this knowledge to specify the chemical reactants of a given product.
We also follow this approach and demonstrate its capability using the retrosynthesis section of the Mol-Instructions~\citep{fang2024mol} dataset, which originates from the USPTO\_500MT~\citep{lu2022unified}. Each sample consists of a product molecule and the required reactant molecules, separated by a period (‘.’). 
For each test sample, we provide the product's SMILES representation, its molecular graph, and the given instruction prompt as input to the model to generate possible reactants in SMILES format.

\section{Prompts}
\label{app_sec:prompts}
In this section, we provide input prompts used to evaluate GPT-3.5 and GPT-4 on molecular description generation task~(Table~\ref{tab:example_prompt_cap}), IUPAC name prediction task~(Table~\ref{tab:example_prompt_iupac}), and property question answering task~(Table~\ref{tab:example_prompt_prop}).
We also provide the input prompts for machine-generated instruction-tuning data generation~(Table~\ref{tab:example_prompt_conv}, \ref{tab:example_prompt_conv_iupac}).

\begin{table*}[t!]
\caption{{Example prompt for GPT in molecular description generation task.} The top block shows the instruction part of the prompts. It includes \textcolor{red}{original instruction from the dataset} and \textcolor{blue}{additional instruction added only for in-context learning}. The bottom block shows optional few-shot examples and target SMILES.}
    \label{tab:example_prompt_cap}
\centering
\begin{minipage}{1.0\columnwidth}\vspace{0mm}    \centering
\begin{tcolorbox} 
    \centering
      \small
    \begin{tabular}{p{0.97\columnwidth} c}
    You are an expert chemist. Please strictly follow the format, no other information can be provided. Given the molecular SMILES, \textcolor{red}{could you provide a description of this molecule?} \textcolor{blue}{You will be provided with several examples of molecules and their descriptions}.&\\
    \hrulefill & \\
    Molecule SMILES: COc1ccc(-c2cc(=O)c3c(O)c(Oc4ccc(-c5cc(=O)c6c(O)cc(O)cc6o5)cc4)c(OC)cc3&\\o2)cc1&\\
    Molecular Description: The molecule is a natural product found in Selaginella tamariscina, Taxodium distichum, and other organisms with data available.&\\
    \\
    ...
    \\
    &\\
    Molecule SMILES: CC=Cc1ccc2oc(-c3cccc(Oc4cc(-c5oc6ccc(C=CC)cc6c5C)ccc4O)c3)c(C)c2c1&\\
    Molecular Description: The molecule is a natural product found in Piper aequale with data available. &\\
    &\\
    Molecule SMILES: C/C=C/c1ccc2oc(-c3ccc(Oc4cc(-c5oc6ccc(/C=C/C)cc6c5C)ccc4O)cc3)c(C)c2c1&\\
    Molecular Description:
    &
    \end{tabular}
\end{tcolorbox}
\end{minipage}
\end{table*}

\begin{table*}[t!]\centering
\caption{{Example prompt for GPT in IUPAC name prediction task.} The top block shows the instruction part of the prompts. It includes \textcolor{blue}{additional instruction added only for in-context learning}. The bottom block shows optional few-shot examples and target SMILES.}
    \label{tab:example_prompt_iupac}
\begin{minipage}{1.0\columnwidth}\vspace{0mm}    \centering
\begin{tcolorbox} 
    \centering
      \small
    \begin{tabular}{p{0.97\columnwidth} c}
       You are an expert chemist. Please strictly follow the format, no other information can be provided. Given the molecular SMILES, your task is to predict the IUPAC name using your experienced chemical IUPAC name knowledge. \textcolor{blue}{You will be provided with several examples of molecules and their IUPAC names.}&\\
    \hrulefill & \\
    Molecule SMILES: COc1cc([C@H]2COc3cc(O)ccc3C2)ccc1O&\\
    The molecule’s IUPAC name is (3S)-3-(4-hydroxy-3-methoxyphenyl)-3,4-dihydro-2H-chromen-7-ol&\\
    \\
    ...
    \\
    &\\
    Molecule SMILES: COc1c([C@@H]2COc3cc(O)ccc3C2)ccc2c1C=CC(C)(C)O2&\\
    The molecule’s IUPAC name is (3R)-3-(5-methoxy-2,2-dimethylchromen-6-yl)-3,4-dihydro-2H-chromen-7-ol&\\
    &\\
    Molecule SMILES: COC1=CC(=O)C(C2COc3cc(O)ccc3C2)=CC1=O&\\
    The molecule’s IUPAC name is
    &
    \end{tabular}
\end{tcolorbox}
\end{minipage}
\end{table*}
\begin{table*}[t!]\centering
\caption{{Example prompt for GPT in property question answering task.} The top block shows the instruction part of the prompts. It includes \textcolor{red}{original instruction from the dataset} and \textcolor{blue}{additional instruction added only for in-context learning}. The bottom block shows optional few-shot examples and target SMILES.}
    \label{tab:example_prompt_prop}
\begin{minipage}{1.0\columnwidth}\vspace{0mm}    \centering
\begin{tcolorbox} 
    \centering
      \small
    \begin{tabular}{p{0.97\columnwidth} c}
       You are an expert chemist. Please strictly follow the format, no other information can be provided. Given the molecular SMILES, \textcolor{red}{what is the energy separation between the HOMO and LUMO of this molecule?} \textcolor{blue}{You will be provided with several examples of molecules and their HOMO-LUMO gap values.}&\\
    \hrulefill & \\
    Molecule SMILES: COCC12OC3CC1C32 & \\
    Output Value: 0.2967 & \\
    \\
    ...
    \\
    \\
    Molecule SMILES: OCCC12CC3C(O1)C32 & \\
    Output Value: 0.305 & \\
    \\
    Molecule SMILES: CCC1C2OC3C1C23C & \\
    Output Value:
    &
    \end{tabular}
\end{tcolorbox}
\end{minipage}
\end{table*}
\begin{table*}[t!]\centering
\caption{{Input prompt for generating multi-turn conversation data based on SMILES and caption.}}
    \label{tab:example_prompt_conv}
\begin{minipage}{1.0\columnwidth}\vspace{0mm}    \centering
\begin{tcolorbox} 
    \centering
      \small
    \begin{tabular}{p{0.97\columnwidth} c}
You are an AI chemical assistant, and you are seeing a single molecule. What you see is provided with SMILES representation of the molecule and sentences describing the same molecule you are analyzing. Answer all questions as you are seeing the molecule.\\
Ask diverse questions and give corresponding answers.\\Include questions asking about the detailed information of the molecule, including the class, conjugate acid/base, functional groups, chemical role, etc.\\Do not ask any question that cannot be answered confidently.\\

    Molecule SMILES: \{SMILES\} & \\
    Caption: \{CAPTION\} & \\
    Conversation:
    &
    \end{tabular}
\end{tcolorbox}
\end{minipage}
\end{table*}
\begin{table*}[t!]\centering
\caption{{Input prompt for generating multi-turn conversation data based on SMILES, caption, and IUPAC.}}
    \label{tab:example_prompt_conv_iupac}
\begin{minipage}{1.0\columnwidth}\vspace{0mm}    \centering
\begin{tcolorbox} 
    \centering
      \small
    \begin{tabular}{p{0.97\columnwidth} c}
You are an AI chemical assistant, and you are seeing a single molecule. What you see is provided with SMILES representation of the molecule and sentences describing the same molecule you are analyzing. In addition, the IUPAC name of the molecule is given. Answer all questions as you are seeing the molecule.\\
Ask diverse questions and give corresponding answers.\\
Include questions asking about the detailed information of the molecule, including the class, conjugate acid/base, functional groups, chemical role, etc.\\
Do not ask any questions that cannot be answered confidently.

    Molecule SMILES: \{SMILES\} & \\
    Caption: \{CAPTION\} & \\
    IUPAC: \{IUPAC\} & \\
    Conversation:
    &
    \end{tabular}
\end{tcolorbox}
\end{minipage}
\end{table*}

\section{Over-smoothing Problems}
\label{app_sec:oversmoothing}
\begin{figure}[t]
    \centering
    \includegraphics[width=1.0\textwidth]{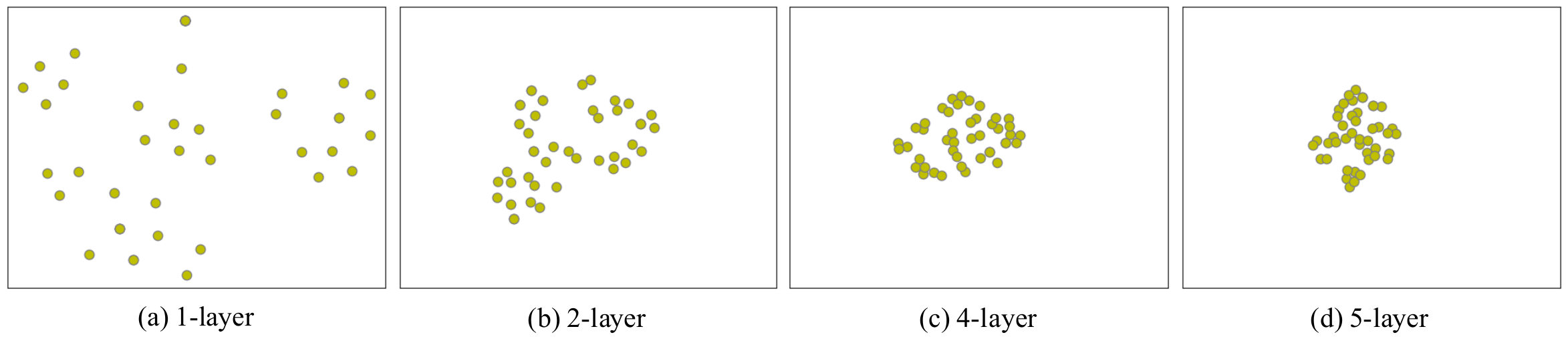}
    \includegraphics[width=1.0\textwidth]{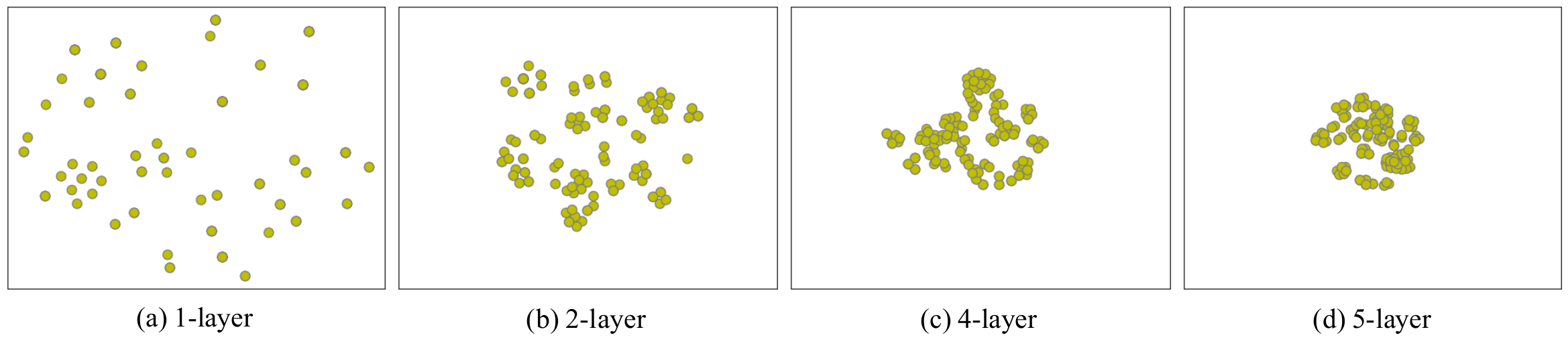}
    \includegraphics[width=1.0\textwidth]{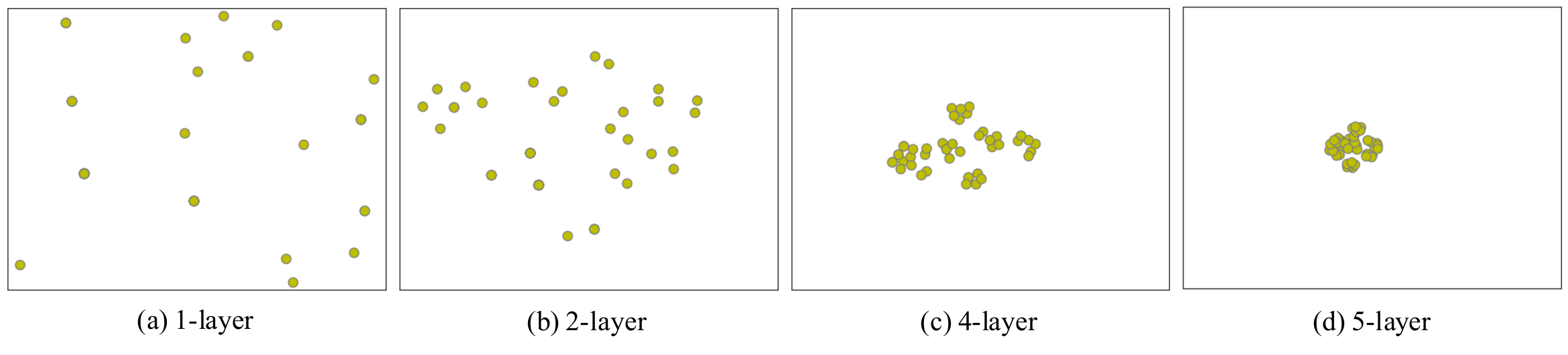}
    \caption{Node representations of graph encoder with 1,2,4,5 layers. As the number of layers increases, node representations collapse.
    } 
    \label{fig_sup:oversmoothing_fig}
\end{figure}
We provide additional samples to show the over-smoothing problems in Figure~\ref{fig_sup:oversmoothing_fig}.
\section{More Qualitative Samples}
\label{app_sec:qual}
\begin{figure}[t]
    \centering
        \includegraphics[width=\textwidth]{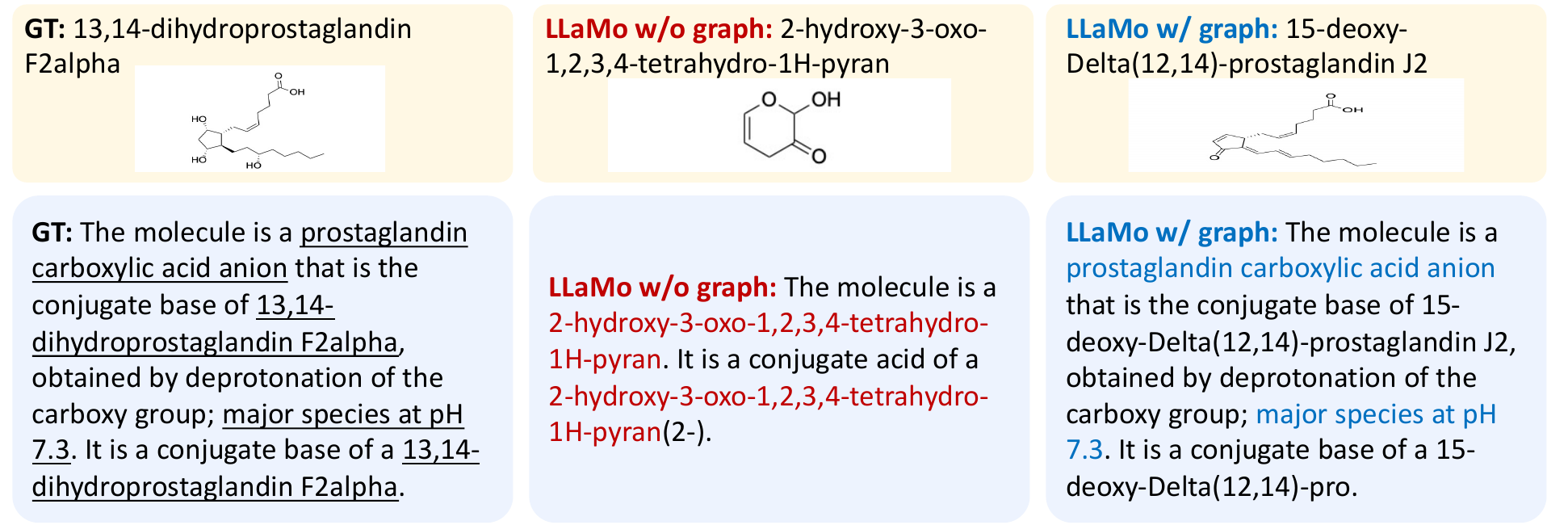}
    \hfill
    \caption{An example of molecular description generation results of LLaMo~w/o graph and LLaMo~w/ graph given the molecule ‘‘CCCCC[C@@H1](CC[C@H1]1[C@@H1](C[C@@H1]([C@@H1]\\1C/C=C$\backslash$CCCC(=O)[O-1])O)O)O)’’.
    }
    \label{fig:qual1_supple}
\end{figure}
\begin{figure}[t]
    \centering
    \begin{subfigure}[t]{0.98\textwidth}
        \includegraphics[width=\textwidth]{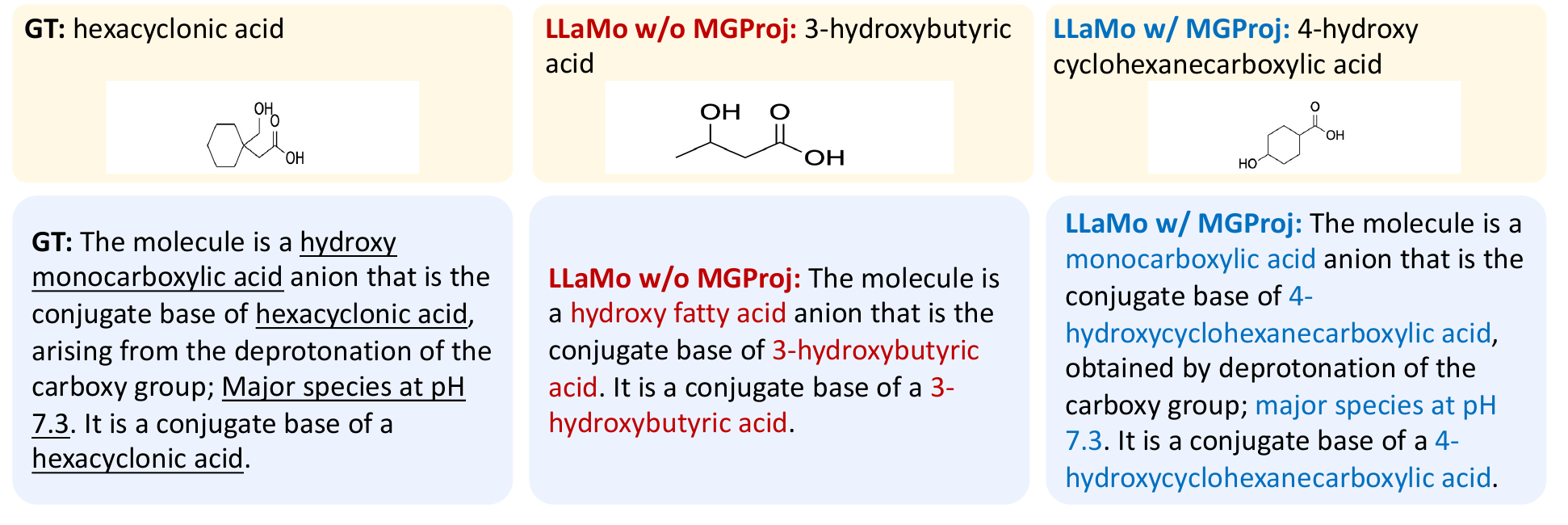}
        \caption{Molecule ‘‘C1CCC(CC1)(CC(=O)[O-1])CO”.}
        \label{subfig:qual2_supple1}
    \end{subfigure}
    \hfill
    \begin{subfigure}[t]{0.98\textwidth}
        \includegraphics[width=\textwidth]{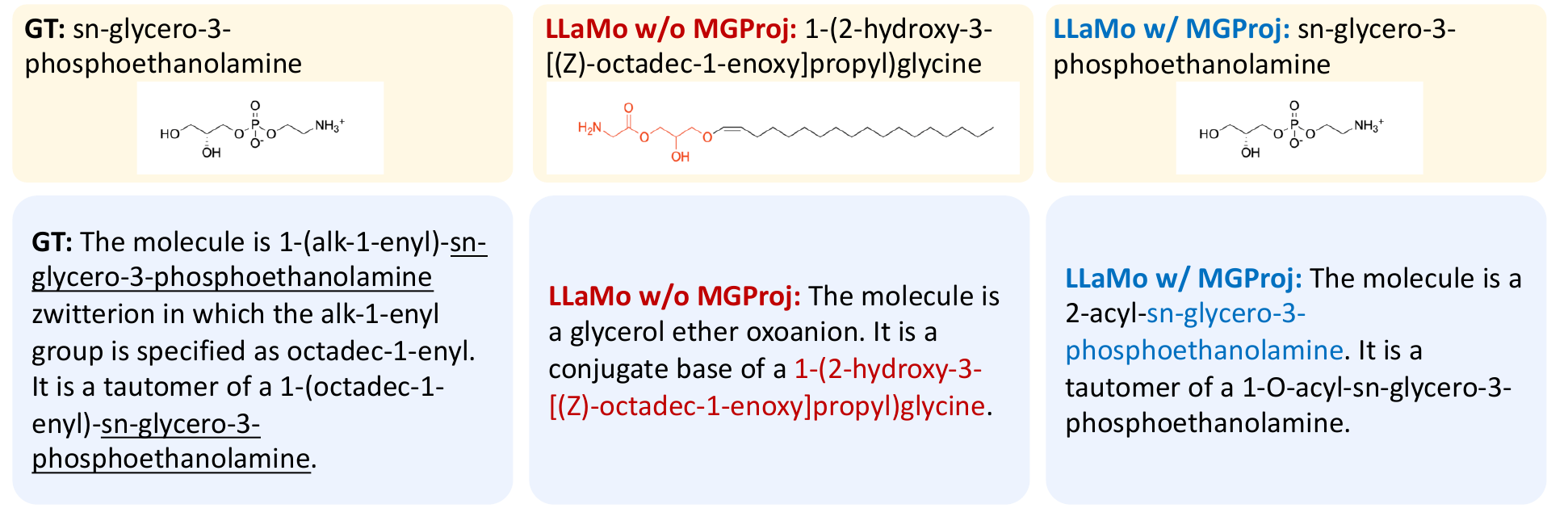}
        \caption{Molecule ‘‘CCCCCCCCCCCCCCCCC=COC[C@H1](COP(=O)([O-1])OCC[NH3+1])O”.
        }
        \label{subfig:qual2_supple2}
    \end{subfigure}
    \caption{Examples of molecular description generation results of LLaMo w/o MGProj and LLaMo w/ MGProj.
    }
    \label{fig:qual2_supple}
\end{figure}

We provide further qualitative results of LLaMo w/ and w/o graph in Figure~\ref{fig:qual1_supple}, and LLaMo w/ and wo MGProj in Figure~\ref{fig:qual2_supple}, respectively.

\clearpage
\section*{NeurIPS Paper Checklist}

\begin{enumerate}

\item {\bf Claims}
    \item[] Question: Do the main claims made in the abstract and introduction accurately reflect the paper's contributions and scope?
    \item[] Answer: \answerYes{} 
    \item[] Justification: We clearly state the main claims in the introduction and abstract.

\item {\bf Limitations}
    \item[] Question: Does the paper discuss the limitations of the work performed by the authors?
    \item[] Answer: \answerYes{} 
    \item[] Justification: Please refer to Appendix~\ref{app_sec:limitations}.

\item {\bf Theory Assumptions and Proofs}
    \item[] Question: For each theoretical result, does the paper provide the full set of assumptions and a complete (and correct) proof?
    \item[] Answer: \answerNA{} 
    \item[] Justification: {Our paper does not include any theoretical analysis.}

    \item {\bf Experimental Result Reproducibility}
    \item[] Question: Does the paper fully disclose all the information needed to reproduce the main experimental results of the paper to the extent that it affects the main claims and/or conclusions of the paper (regardless of whether the code and data are provided or not)?
    \item[] Answer: \answerYes{} 
    \item[] Justification: We provide the implementation details in the Experiments section.

\item {\bf Open access to data and code}
    \item[] Question: Does the paper provide open access to the data and code, with sufficient instructions to faithfully reproduce the main experimental results, as described in supplemental material?
    \item[] Answer: \answerYes{} 
    \item[] Justification: We make our code and dataset publicly available.

\item {\bf Experimental Setting/Details}
    \item[] Question: Does the paper specify all the training and test details (e.g., data splits, hyperparameters, how they were chosen, type of optimizer, etc.) necessary to understand the results?
    \item[] Answer: \answerYes{} 
    \item[] Justification: We provide the implementation details in the Experiments section.

\item {\bf Experiment Statistical Significance}
    \item[] Question: Does the paper report error bars suitably and correctly defined or other appropriate information about the statistical significance of the experiments?
    \item[] Answer: \answerNA{} 
    \item[] Justification: We report the performances with single-run experiments.

\item {\bf Experiments Compute Resources}
    \item[] Question: For each experiment, does the paper provide sufficient information on the computer resources (type of compute workers, memory, time of execution) needed to reproduce the experiments?
    \item[] Answer: \answerYes{} 
    \item[] Justification: We provide the implementation details in the Experiments section.
    
\item {\bf Code Of Ethics}
    \item[] Question: Does the research conducted in the paper conform, in every respect, with the NeurIPS Code of Ethics \url{https://neurips.cc/public/EthicsGuidelines}?
    \item[] Answer: \answerYes{} 
    \item[] Justification: We follow the NeurIPS Code of Ethics in our research.

\item {\bf Broader Impacts}
    \item[] Question: Does the paper discuss both potential positive societal impacts and negative societal impacts of the work performed?
    \item[] Answer: \answerYes{} 
    \item[] Justification: We discuss both positive and negative societal impacts in Appendix~\ref{app_sec:broader}.
    
\item {\bf Safeguards}
    \item[] Question: Does the paper describe safeguards that have been put in place for responsible release of data or models that have a high risk for misuse (e.g., pretrained language models, image generators, or scraped datasets)?
    \item[] Answer: \answerYes{} 
    \item[] Justification: We release the code and checkpoint with the safeguards.

\item {\bf Licenses for existing assets}
    \item[] Question: Are the creators or original owners of assets (e.g., code, data, models), used in the paper, properly credited and are the license and terms of use explicitly mentioned and properly respected?
    \item[] Answer: \answerYes{} 
    \item[] Justification: We explicitly mention the license of used assets in the supplement.

\item {\bf New Assets}
    \item[] Question: Are new assets introduced in the paper well documented and is the documentation provided alongside the assets?
    \item[] Answer: \answerYes{} 
    \item[] Justification: We explain the data generation process in Section 4 and make the data publicly available.

\item {\bf Crowdsourcing and Research with Human Subjects}
    \item[] Question: For crowdsourcing experiments and research with human subjects, does the paper include the full text of instructions given to participants and screenshots, if applicable, as well as details about compensation (if any)? 
    \item[] Answer: \answerNA{} 
    \item[] Justification: We do not use crowdsourcing in this paper.

\item {\bf Institutional Review Board (IRB) Approvals or Equivalent for Research with Human Subjects}
    \item[] Question: Does the paper describe potential risks incurred by study participants, whether such risks were disclosed to the subjects, and whether Institutional Review Board (IRB) approvals (or an equivalent approval/review based on the requirements of your country or institution) were obtained?
    \item[] Answer: \answerNA{} 
    \item[] Justification: We do not use crowdsourcing in this paper.

\end{enumerate}

\end{document}